\documentclass{article}
\usepackage{ijcai24}

\usepackage{times}
\usepackage{soul}
\usepackage{url}
\usepackage[hidelinks]{hyperref}
\usepackage[utf8]{inputenc}
\usepackage[small]{caption}
\usepackage{graphicx}
\usepackage{amsmath}
\usepackage{amsthm}
\usepackage{amssymb}
\usepackage{booktabs}
\usepackage{algorithm}
\usepackage{algorithmic}
\usepackage[switch]{lineno}
\usepackage{multirow}
\usepackage{xcolor}         % colors

\title{Prompt Learning for Generalized Vehicle Routing}

\author{
Fei Liu$^1$,
Xi Lin$^1$,
Weiduo Liao$^1$,
Zhenkun Wang$^{2,}$\thanks{corresponding author},
Qingfu Zhang$^{1,*}$, 
Xialiang Tong$^3$, \And
Mingxuan Yuan$^3$
\\
\affiliations
$^1$City University of Hong Kong\\
$^2$Southern University of Science and Technology\\
$^3$Huawei Noah’s Ark Lab\\
\emails
\{fliu36-c,xi.lin,weiduliao2-c\}@my.cityu.edu.hk,
wangzk3@sustech.edu.cn,
qingfu.zhang@cityu.edu.hk,
\{tongxialiang,yuan.mingxuan\}@huawei.com
}

\begin{document}

\maketitle

\begin{abstract}

Neural combinatorial optimization (NCO) is a promising learning-based approach to solving various vehicle routing problems without much manual algorithm design. However, the current NCO methods mainly focus on the in-distribution performance, while the real-world problem instances usually come from different distributions. A costly fine-tuning approach or generalized model retraining from scratch could be needed to tackle the out-of-distribution instances. Unlike the existing methods, this work investigates an efficient prompt learning approach in NCO for cross-distribution adaptation. To be concrete, we propose a novel prompt learning method to facilitate fast zero-shot adaptation of a pre-trained model to solve routing problem instances from different distributions. The proposed model learns a set of prompts among various distributions and then selects the best-matched one to prompt a pre-trained attention model for each problem instance. Extensive experiments show that the proposed prompt learning approach facilitates the fast adaptation of pre-trained routing models. It also outperforms existing generalized models on both in-distribution prediction and zero-shot generalization to a diverse set of new tasks. Our code implementation is available online\footnote{https://github.com/FeiLiu36/PromptVRP}.

\end{abstract}

\section{Introduction}

The Vehicle Routing Problem (VRP) can be found in many real-world applications such as logistics, transportation, retail distribution, waste collection, and manufacturing~\cite{toth2014vehicle}. Its objective is to manage a fleet of vehicles optimally, minimizing the total cost while satisfying the demands of customers. As an NP-hard problem, exact methods can hardly solve it efficiently, while heuristic algorithms require costly handcrafted designs with domain knowledge. In contrast, neural combinatorial optimization (NCO), which learns a heuristic based on neural networks, has received growing attention~\cite{bengio2021machine,raza2022vehicle,bai2023analytics,bogyrbayeva2024machine} due to its potential ability to generate high-quality solutions without much human effort~\cite{vinyals2015pointer,kool2018attention,bogyrbayeva2024machine}.

Most existing neural combinatorial optimization methods focus on solving in-distribution instances, while real-world routing problem instances are typically from diverse distributions. Therefore, their performance could deteriorate dramatically on out-of-distribution instances~\cite{bi2022learning,zhou2023towards}. Recent efforts have focused on enhancing the generalization capabilities for out-of-distribution tasks~\cite{jiang2022learning,bi2022learning,fu2021generalize,pan2023h-tsp,manchanda2023generalization,drakulic2023bq,jiang2023multi,zhou2023towards}. The majority of these approaches involve training a single generalized model using meta-learning techniques~\cite{jiang2022learning,bi2022learning,manchanda2023generalization,zhou2023towards}, which can be adapted effectively to tackle instances from different distributions. However, these methods often necessitate complex and time-intensive meta-learning-based training, while the learning capacity is constrained by the fixed model structure.

\begin{figure*}[t]
    \centering
    \includegraphics[width=0.8\linewidth]{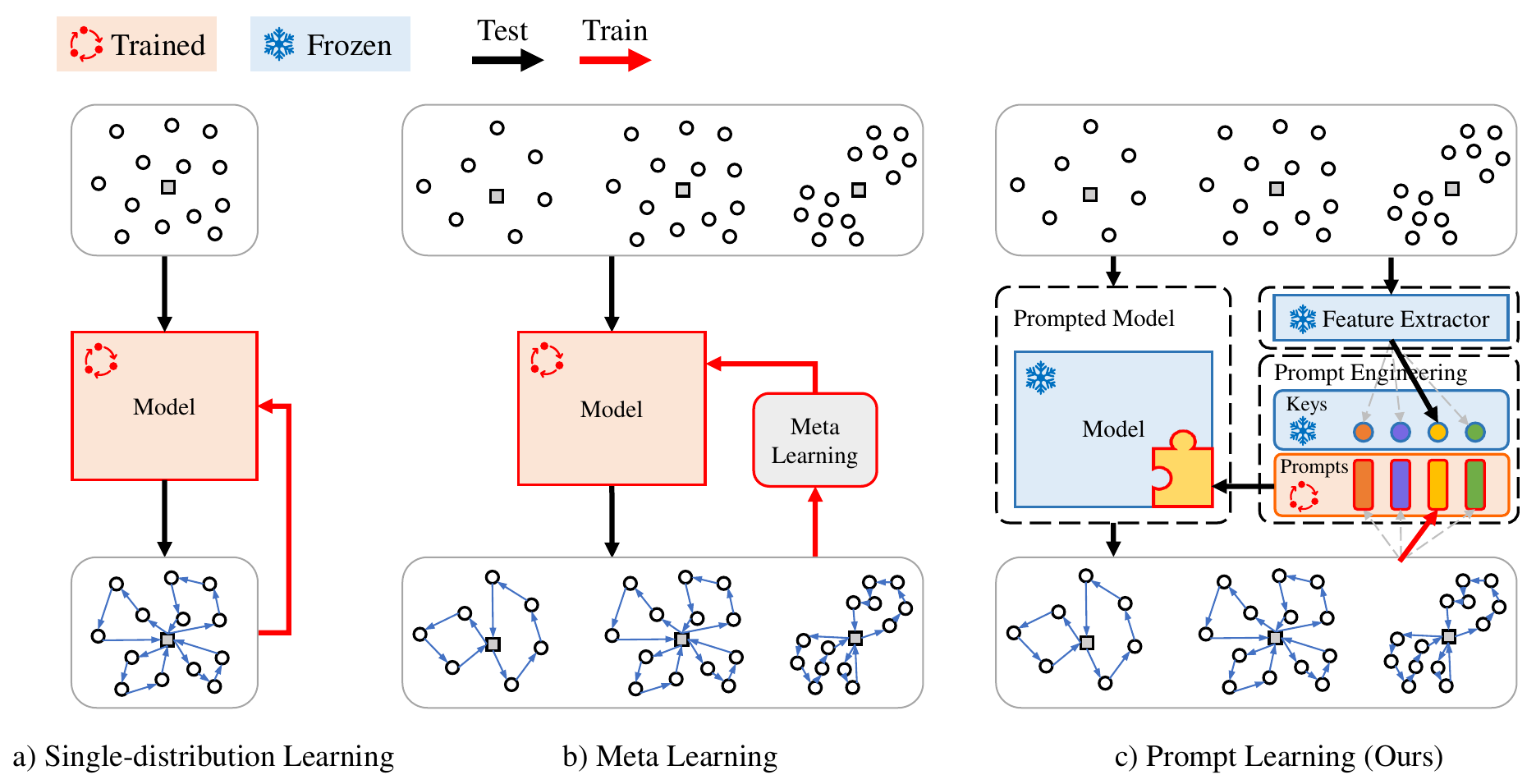}
    \caption{Three approaches for cross-distribution neural combinatorial optimization. \textbf{a) Single-distribution Learning:} Single-distribution learning focuses on solving problem instances from the same distribution, and hence its performance usually significantly deteriorates for out-of-distribution cases. \textbf{b) Meta Learning:} Meta learning builds a single model to handle problem instances from diverse distributions. It requires a complicated and time-consuming training strategy, while the learning capacity might be limited by the static model structure. \textbf{c) Prompt Learning (Ours):} The proposed prompt learning incorporates a trainable key-prompt pool into a frozen NCO model to tackle different problem instances across diverse distributions. For inference, it can automatically select the most suitable prompt for a given instance, and adjust the prompt-based attention in a zero-shot manner to obtain better performance.}
    %\caption{Three frameworks for constructive NCO: a) conventional \textbf{single-distribution learning}, b) \textbf{meta learning} for cross-distribution performance, and c) our proposed \textbf{prompt learning}. The single-distribution learning focuses on one distribution. Its performance usually deteriorates in out-of-distribution cases. The second meta learning approach trains a generalized model over multiple distributions. It has much robust cross-distribution performance. In contrast to these existing frameworks, which learn over model parameters, our proposed prompt learning learns a set of prompts to adapt a fixed pre-trained model across multiple distributions as well as zero-shot generalize to new distributions. }
    \label{fig_framework}
\end{figure*}

This paper proposes a novel approach, which uses prompt learning~\cite{zhou2022learning,liu2023pre} to enable fast zero-shot adaptation of a pre-trained NCO model to tackle out-of-distribution routing problem instances. As shown in Figure~\ref{fig_framework}, we keep the pre-trained encoder-decoder NCO model fixed and efficiently learn a pool of key-prompt pairs incorporated into the model for handling different problem instances from diverse distributions. The cross-distribution information is learned through the shared prompts. For solving a new problem instance, the most suitable key will be automatically selected, and its matched prompt will be used to adjust the pre-trained NCO model in a zero-shot manner for better inference. In this way, the proposed prompt learning method can efficiently extend the learning capacity of the pre-trained NCO model, demonstrating competitive generalization performance.

The contributions of this work are summarized as follows:

%\xl{For the first two bullet points, we should emphasize the effect/significance of our the contribution (why it is good), rather than the technical details.} \fei{revised}
\begin{itemize}

    % \item We are the first to study prompt learning for solving cross-distribution vehicle routing problems. Instead of direct train the model on multiple tasks with meta learning, we propose to learn a set of prompts, which are then used to prompt a pre-trained model for vehicle routing with various distributions.
        
    % Instead of training the model on multiple distributions with meta learning, we propose to learn a set of prompts, which are then used to prompt a pre-trained model for vehicle routing with various distributions.

   % \item We test our proposed method on various cross-distribution routing problems as well as benchmark instances. Our method significantly reduces the training cost and outperforms the existing meta learning methods in terms of cross-distribution performance.

    %\item We show how to do prompt learning in combinatorial optimization and propose the first prompt learning method for cross-distribution vehicle routing problems.

    % \item Instead of training a generalized model through meta learning, we learn a set of prompts to adapt a fixed pre-trained model to diverse distributions. The capacity of the model is further extended while the training cost is reduced.

    \item We investigate how to incorporate prompt learning into neural combinatorial optimization and propose the first prompt learning method for solving cross-distribution vehicle routing problems. 
    
    \item We develop a novel and efficient prompt-based attention model to tackle different routing problem instances from diverse distributions via fast zero-shot adaption.  
    
   \item We evaluate our proposed prompt learning method on extensive cross-distribution routing instances as well as benchmark instances. With a much lower training cost, our method achieves superior performance compared to existing meta learning methods.

\end{itemize}

\section{Related Works}

% NCO

%\paragraph{Neural Combinatorial Optimization (NCO) }
\subsection{Neural Combinatorial Optimization (NCO)}

NCO intends to automatically learn a heuristic based on neural networks for solving combinatorial optimization problems. Compared to the other approaches, such as exact methods and heuristic algorithms, it usually generate high-quality solutions with a fast runtime~\cite{bengio2021machine}. As a result, NCO has gained much attention in the past decade~\cite{bengio2021machine,bogyrbayeva2024machine}. As one of the most important combinatorial optimization problems, the vehicle routing problems have been extensively studied in NCO works~\cite{vinyals2015pointer,bello2016neural,nazari2018reinforcement,kool2018attention,li2022overview}.

There are mainly two categories of works along this line:  the end-to-end construction-based methods~\cite{vinyals2015pointer,bello2016neural,kool2018attention,kwon2020pomo,joshi2022learning} and the improvement-based methods~\cite{chen2019learning,hottung2019neural,chen2019learning,kool2022DPDP}. The former aims to construct a solution without any assistance from classical solvers, while the latter incorporates additional algorithms to improve performance. This work focuses on the construction-based method. 

% NCO for Generalization performance

%\paragraph{NCO for Cross-distribution Routing Problem}
\subsection{NCO for Cross-distribution Routing Problem}

Several meta learning methods have been developed to improve the out-of-distribution generalization performance for routing problems. Jiang~\shortcite{jiang2022learning} and Bi~\shortcite{bi2022learning} explored the robust optimization over multiple geometrical distributions. Several works~\cite{fu2021generalize,pan2023h-tsp,manchanda2023generalization,drakulic2023bq} studied the generalization to large-scale problems. Zhou~\shortcite{zhou2023towards} considered generalization in terms of both problem size and geometrical distribution. Most of the existing works adopt a single generalized model and use meta learning methods to improve cross-distribution performance, which might lead to time-consuming training and constrained learning capacity. 

% In contrast, we propose to use prompt learning for cross-distribution vehicle routing and zero-shot generalization.

% Prompt learning 

%\paragraph{Prompt Learning}
\subsection{Prompt Learning}

Prompt learning has recently gained significant attention in many research areas, such as natural language processing (NLP)~\cite{liu2023pre}, computer vision (CV)~\cite{jia2022visual,zhou2022learning,ge2023domain}, and reinforcement learning (RL)~\cite{xu2022prompting}. In NLP, seminal works like GPT-3~\cite{brown2020language} and InstructGPT~\cite{ouyang2022training} showcase the effectiveness of prompts in guiding text generation for diverse tasks. In CV, prompt learning can enable few-shot learning~\cite{zhang2023prompt} and improves image captioning~\cite{wang2023controllable} by conditioning on specific instructions. In RL, prompt learning can leverage the flexible adaption of prompts to enhance the few-shot policy generalization performance~\cite{xu2022prompting}. 

In recent years, many well-trained models have been developed for combinatorial optimization~\cite{kool2018attention,kwon2020pomo,bogyrbayeva2024machine}. However, the effective utilization of these pre-trained models has not been thoroughly investigated. This paper proposes a prompt learning method to efficiently adapt a fixed pre-trained model for addressing cross-distribution vehicle routing problems.

%Although there are many well-trained pre-trained models developed in recent years for combinatorial optimization, the efficient adaptation of these models has not been well explored. This paper, for the first time, proposes a prompt learning method for solving cross-distribution vehicle routing problems.    

% Prompt learning has recently gained significant attention~\cite{liu2023pre}. The applications can be found in a range of domains, including natural language processing (NLP)~\cite{liu2023pre}, computer vision (CV)~\cite{jia2022visual,zhou2022learning}, and more. In NLP, works like GPT-3~\cite{brown2020language} and InstructGPT~\cite{ouyang2022training} showcase the effectiveness of prompts in guiding text generation for diverse tasks. In CV, prompt learning enables few-shot learning~\cite{zhang2023prompt} and improves image captioning~\cite{wang2023controllable} by conditioning on specific instructions. Other areas, such as reinforcement learning, have also leveraged prompts to enhance agent behavior. This paper, for the first time, studies how to adopt prompt learning for solving combinatorial optimization problems. 

% ???????????
% no section no

% ?????????????
% make it clear the following is your work
% ????? 

\section{Prompt Learning for Routing}

% ????
% add more subsection

% explain main idea ?????
% why novel
% ??????

% \xl{

% Before directly jumping into the proposed framework, we might need a brief background subsection to include: 

% \begin{itemize}
%     \item The Problem to Study: a formal formulation of the NCO4VRP with diverse distributions;
%     \item The Current Method: the key formulation for meta learning based NCO (or a figure to show their model structure), to emphasize the shortcoming of this approach;
% \end{itemize}

% }

\subsection{Problem Formulation}

We denote a basic capacited vehicle routing problem (CVRP) on an undirected graph $G = (V,E)$. $V=\{v_0,\dots,v_n\}$, where $v_0$ is the depot and $v_1,\dots,v_n$ are the $n$ customers. $V_c=\{v_1,\dots,v_n\}$ is the customer set. For the $i$-th customer, there is a demand $d_i$. $E=\{e_{ij}\},i,j\in \{1,\dots,n\}$ are the edges between every two nodes. For each edge $e_{ij}$, there is an associated cost (distance) $c_{ij}$. A fleet of homogeneous vehicles with a capacity of $C$ is sent out from the depot to visit the customers and return to the depot. All the customer's demands should be served. Each customer must be visited once. The objective is to minimize the total traveling distance of all the routes with all the constraints satisfied.

\subsection{Main Idea and Basic Framework}

The typical constructive-based NCO methods~\cite{kool2018attention,kwon2020pomo} use an attention-based encoder-decoder model to directly construct a valid solution (e.g., a tour) for the mentioned routing problem. They learn the best model parameters for the attention model to minimize the total distances. In this case, the objective of model training would be:
\begin{equation}
    \theta^*=\arg \min _{\theta} \mathbb{E}_{\mathcal{G} \sim p(\mathcal{\mathcal{G}})} \mathcal{L}\left(\tau \mid \theta, \mathcal{G} \right)
\end{equation}
where $\mathcal{G}$ represents the given instance, $\theta$ is the model parameter, and $\tau$ is the trajectory (e.g., tour) constructed by the model. The goal is to find the best model parameter $\theta^*$ to minimize the average total distance (as the training loss $\mathcal{L}$) for $\tau$ over a given distribution $p(\mathcal{G})$.

When explicitly considering multiple distributions in model training, most existing works treat each distribution as a task and use meta learning for model training~\cite{manchanda2023generalization,zhou2023towards}. The objective is to learn a single yet robust model parameter $\theta^*$ that can generalize well over various distributions.:
\begin{equation}
    \theta^*=\arg \min _{\theta} \frac{1}{T} \sum_{i=1}^T \mathbb{E}_{\mathcal{G} \sim p_i(\mathcal{\mathcal{G}})} \mathcal{L}\left(\tau \mid \theta, \mathcal{G} \right)
\end{equation}
where $T$ is the number of tasks, and $p(\mathcal{G})$ represents the distribution over $i$-th task.

Different from the two approaches, we propose to incorporate prompt learning into the NCO model for tackling cross-distribution vehicle routing problems. The objective can be formulated as:
\begin{equation}
\begin{aligned}
    &\{P^*_1,\dots,P^*_M\} \\ 
    &= \arg\min _{\{P_1,\dots,P_M\}} \frac{1}{T} \sum_{i=1}^T \mathbb{E}_{\mathcal{G} \sim p_i(\mathcal{\mathcal{G}})} \mathcal{L}\left(\tau \mid P, \theta, \mathcal{G} \right)
\end{aligned}
\end{equation}
where $\{P_1,\dots,P_M\}$ are $M$ prompts, and $P$ is the selected prompt for each given instance. In this prompt-based model, we can learn the $M$ prompts instead of the entire set of model parameters $\theta$. The objective here is to learn the best prompts that adapt the pre-trained model with a fixed $\theta$ for across-distribution performance.

% \xl{

% Is there a way to use clear formulation to introduce the general idea of prompt learning or even prompt learning for NCO?

% }

% The main idea of this paper is to learn a set of prompts instead of training the entire model on routing instances with various distributions. We are the first to study prompt learning for cross-distribution vehicle routing problems. As shown in Figure~\ref{fig_framework}, our proposed prompt learning framework is built on the pre-trained networks and the only learnable component is the prompts. The feature extractor is used to transfer an input instance into a feature vector, and then we can identify the best-matched key from the key-prompt pair pool to match the input feature. The associated prompt of the best-matched key will be used to prompt the pre-trained neural solver and a solution will be generated by the solver.

% The novelty and advantages of our method are threefold: 1) Prompt learning can be much faster than directly training the model. 2) Different prompts are learned simultaneously for different distributions. The generalization performance of a single model is significantly extended. 

As illustrated in Figure~\ref{fig_framework}, our proposed prompt learning consists of three main components: 1) feature extractor, 2) prompt engineering, and 3) prompted model. We adopt pre-trained attention networks as the feature extractor and the model, which remain fixed during training and testing. The keys are also predetermined based on the features of the randomly generated training instances. The only adjustable components are the prompts. The input instance is fed into both the model and the feature extractor. The feature extractor converts the input instance into a feature vector, allowing us to identify the most appropriate key from the key-prompt pair pool to match the input feature. The key and prompt are coupled together. The associated prompt of the best-matched key is then used to prompt the pre-trained model. A solution is generated by the prompted model, based on the selected prompt. The solution is used to calculate rewards for updating the selected prompt during training.

\subsection{Feature Extractor}

In this work, we directly use the encoder of the attention model~\cite{kool2018attention} as the feature extractor. The encoder consists of $L$ stacked multi-head attention (MHA) blocks. The input of the encoder is the node features $f_i,i=1,\dots,n$. In our experiments, the input features for the $i$-th node are denoted as $f_i = \{x_i, y_i, d_i\}$, where $(x_i, y_i)$ are the coordinates and $d_i$ is the demand. The input features are embedded through a linear projection to generate the initial feature embedding $h_i^{(0)}$. The skip connections~\cite{he2016deep} and instance normalization (IN) are used in each MHA layer: 
\begin{equation}
    \begin{gathered}
        \hat{h}_i^{(l)}=IN^l\left(h_i^{(l-1)}+MHA_i^l\left(h_1^{(l-1)}, \ldots, h_n^{(l-1)}\right)\right), \\ 
        h_i^{(l)}=IN^l\left(\hat{h}_i+F F^l\left(\hat{h}_i\right)\right),
    \end{gathered}
\end{equation}
where $l$ and $l-1$ represent the current and last embedding layers, respectively. The feedforward (FF) layer contains a hidden sublayer with ReLU activations. The above encoding process generates the final node embeddings $h_i^{(L)}$. %This encoding is performed only once, and the static node embeddings are reused for every decoding step.

Different from the commonly used feature extraction approach in CV and NLP, which uses the embedding of a specific hidden layer, we concatenate the embeddings from multiple layers.  Specifically, we concatenate the output layer of every MHA (before normalization): 
\begin{equation}
    \begin{gathered}
        F^l = \frac{1}{n} \sum_{i=1}^n {\left( \hat{h}_i^{(l-1)} + MHA^l\left(\hat{h}_i^{(l-1)}\right) \right)}, \\
        F = cat\{ F^1 ,F^2, \dots,F^L  \},
    \end{gathered}
\end{equation}
where $F^l$ is the hidden embedding before the last norm layer of the $l$-th MHA and $F$ is the concatenated feature of all $L$ layers. Each hidden embedding $F^l$ is averaged over all $n$ nodes to facilitate generalization across different problem sizes. The final output feature for prompt engineering is adjusted by standard scalarization, given as $F = (F - mean) / stand$, where $mean$ and $stand$ represent the mean and standard deviation of the preprocessing instances, respectively. These preprocessing instances are employed for determining the keys. The mean and standard deviation are calculated element-wise.

\subsection{Prompt Engineering}

We maintain a key-prompt pair pool, which consists of $M$ key-prompt pairs $\{K_i, P_i\},i=1,\dots,M$, where $K_i$ and $P_i$ are the $i$-th key and prompt, respectively. Each pair has a fixed key and a learnable prompt. For each input feature $F_i$, we find the best-matched key $\hat{K}=\min S(F_i,K_j),j=1,\dots,M$, where $S()$ is the distance function. The distance function we employ is the Euclidean distance of the input feature and the key. The prompt $\hat{P}$ associated with $\hat{K}$ is then selected to prompt the pre-trained neural solver. In each batch with $B$ instances, $B$ keys are chosen, and the associated $B$ prompts are updated during training.

The keys $K_i,i=1,\dots, M$ are predetermined vectors of the same size as the feature. They remain fixed throughout the training process. We randomly sample $128$ instances from each of the $341$ distributions, resulting in a total of $43,648$ instances for generating the feature data. The $341$ distributions are introduced in the Appendix. For each instance $i$, we utilize the feature extractor introduced in equation (5) to extract the features $F_i$. We divide the samples into four groups based on problem sizes. For each group, we employ K-means clustering to cluster the samples into $N$ clusters. The cluster centers of the features are then used as the keys. Ultimately, we obtain $M = 4 \cdot N$ vector cluster centers, which are utilized as the keys for prompt learning.

For each key $K_i$, we randomly initialize a vector as the associated prompt $P_i$ according to a uniform distribution and scale the prompt within the range $[-1,1]$.

The key-prompt pairs are connected only in terms of utilization, meaning the associated prompts are used based on selected keys. While their values are decoupled, we only update prompts with key fixed during training. The structure is intentionally kept simple, without dynamically adjusting both keys and prompts. Furthermore, the sizes of the keys and pairs are also different. The former is determined by the feature size, while the latter should be sufficiently long to prompt the pre-trained model, which is introduced in the next subsection.

\begin{table*}[tbp]
\centering
\resizebox{0.8\textwidth}{!}{%
\begin{tabular}{llccccccccc}
\toprule
\multicolumn{1}{c}{\multirow{2}{*}{Method}} & \multicolumn{1}{c}{\multirow{2}{*}{Training Cost}} & \multicolumn{3}{c}{50 } & \multicolumn{3}{c}{100 } & \multicolumn{3}{c}{200} \\

\multicolumn{1}{c}{}& \multicolumn{1}{c}{}       & Dis.   & Gap     & Time  & Dis.   & Gap     & Time   & Dis.   & Gap     & Time   \\
\midrule
HGS     & /  & 10.37  & 0.00\%  & 1.4 h & 15.48  & 0.00\%  & 2.8 h  & 21.87  & 0.00\%  & 5.6 h  \\
LKH3    & /  & 10.42  & 0.49\%  & 1.4 h & 15.59  & 0.69\%  & 2.8 h  & 22.89  & 4.69\%  & 16.7 h \\
\midrule
POMO& /  & 10.98  & 5.92\%  & 1.5 s & 15.82  & 2.18\%  & 2.7 s  & 23.27  & 6.41\%  & 17 s   \\
Meta POMO & \textgreater{}3 d& 10.77  & 3.89\%  & 1.5 s & 16.15  & 4.28\%  & 2.9 s  & 23.14  & 5.83\%  & 18 s   \\
Omni    & 3 d& 10.99  & 5.98\%  & 1.5 s & 16.04  & 3.58\%  & 2.9 s  & 22.80  & 4.29\%  & 18 s   \\
Prompt  & 1 d& 10.70  & 3.20\%  & 1.5 s & 15.88  & 2.57\%  & 2.9 s  & 22.65  & 3.58\%  & 18 s   \\
Prompt top-8         & 1 d& \textbf{10.63}  & \textbf{2.51\%}  & 12 s  & \textbf{15.78}  &\textbf{ 1.94\% } & 23 s   & \textbf{22.58}  & \textbf{3.25\% } & 2.4 m  \\
\midrule
POMO aug      & /  & 10.72  & 3.40\%  & 5 s   & 15.69  & 1.36\%  & 16 s   & 23.00  & 5.19\%  & 86 s   \\
Meta POMO aug       & \textgreater{}3 d& 10.60  & 2.22\%  & 5 s   & 15.96  & 3.08\%  & 16 s   & 22.90  & 4.75\%  & 88 s   \\
Omni aug& 3 d& 10.75  & 3.69\%  & 5 s   & 15.86  & 2.43\%  & 16 s   & 22.63  & 3.51\%  & 88 s   \\
Prompt aug& 1 d& 10.54  & 1.67\%  & 5 s   & 15.74  & 1.65\%  & 16 s   & 22.46  & 2.74\%  & 89 s   \\
Prompt top-8 aug     & 1 d& \textbf{10.51}  & \textbf{1.31\%}  & 40 s  & \textbf{15.68}  & \textbf{1.26\%}  & 2.1 m  & \textbf{22.43}  & \textbf{2.56\% } & 12 m \\
\bottomrule
\end{tabular}%
}
\caption{Comparison of different methods on three training distributions.} ~\label{fig:train}
\end{table*}

\subsection{Prompted Model}

We choose the well-known Attention model~\cite{kool2018attention,kwon2020pomo} as our pre-trained model because it is extensively employed in various routing problems~\cite{bogyrbayeva2024machine}. The model consists of a six-layer encoder and a one-layer decoder. During inference, the encoder inferences once, and the solution of the target routing instance is generated iteratively by the decoder. The selected prompts are used for prompting the six-layer encoder, which allows more control over the pre-trained attention model.

\paragraph{Encoder} The structure of the pre-trained encoder is the same as that used for the feature extractor. It consists of a six-layer MHA, with the linear projection $h_i^{(0)}$ of instance feature $f_i$ as the input and the final node embedding $h_i^{(L)}$ as the output. 

\paragraph{Prompted Encoder}
The selected prompt $P$ from prompt engineering is firstly split into $L$ subprompts $P^{l},l=1,\dots, L$. Each subprompt $P^{l}$ is used for the corresponding embedding layer of the pre-trained encoder. $P^{l}$ has a length of $D\cdot E$, where $D$ is the number of tokens and $E$ is the length of the token. Then, the $l$-th subprompt $P^{l}$ is reshaped into $D$ prompt tokens $p_i^{(l)},i=1,\dots,D$:

\begin{equation} 
    \begin{aligned}
        P & =  \{P^1,\dots,P^L\} \\
        & =  \{p_1^{(1)},\dots,p_D^{(1)}, \dots, p_1^{(L)},\dots,p_D^{(L)} \}.
    \end{aligned}
\end{equation}

These tokens are concatenated into the corresponding $l$-th layer of the encoder. Specifically, for the $l$-th MHA, the $D$ prompt tokens are concatenated with the input hidden layer as follows:
\begin{equation}
    \begin{aligned}
        &\hat{h}_i^{(l)}=I N^l \\
        &\left(h_i^{(l-1)}+M H A_i^{(l)}\left(h_1^{(l-1)}, \ldots, h_n^{(l-1)}, \overbrace{p_{1}^{(l)}, \dots, p_{D}^{(l)}}^{\text{D prompt tokens}}\right)\right), \\
        % F F\left(\hat{h}_i^{(l)}\right)=W_1^F \operatorname{ReLU}\left(W_0^F \hat{h}_i^{(l)}+b_0^F\right)+b_1^F \\ 
        &h_i^{(l)}=I N^l\left(\hat{h}_i+F F^l\left(\hat{h}_i\right)\right).
    \end{aligned}
\end{equation}

As a result, the length of the input tokens of $l$-th layer of MHA is always larger than the input tokens of $l-1$-th layer by $D$. There will be $n + L\cdot D$ output embedding tokens in the last layer of the encoder. We only use the first output $n$ embedding tokens for the decoder instead of all the $n + L\cdot D$ tokens. The first $n$ embedding tokens represent the $n$ nodes of the instance, which are controlled by the $L\cdot D$ prompt tokens.

\paragraph{Decoder} The decoder constructs a solution sequentially. It consists of one MHA layer and one single-head attention (SHA) layer with clipping. The MHA is slightly different from that used in the encoder, where the skip connection, instance normalization, and FF sublayer are now not used~\cite{kool2018attention}. The $t$-th step of inference is as follows:

\begin{equation}
    \begin{aligned}
        &\hat{h}_c = MHA_c \left( h_1^{(L)}, \ldots, h_n^{(L)}, h_t^{(L)}, a_t \right),\\
        &u_1 \dots, u_n = SHA_c \left( h_1^{(L)}, \ldots, h_n^{(L)}, \hat{h}_c \right),
    \end{aligned}
\end{equation}
where $h_t^{(L)}$ and $a_t$ represent the embedding of selected node and attribute in the $t$-th step, respectively. The output embedding of MHA $\hat{h}_c$ is used as the input of the SHA. The SHA outputs the probabilities of choosing the next node using a softmax $p_i =\frac{e^{u_{ i}}}{\sum_j e^{u_{j}}}$ with the unsatisfied nodes masked. We omit the step indicator $t$ for readability. The detailed structure of the MHA and SHA can be found in Kwon~\shortcite{kwon2020pomo}.

\subsection{Training with Reinforcement Learning}

%\xl{can we make it specific for prompt learning?}\fei{}
 
We use the REINFORCE algorithm with a shared baseline following Kwon~\shortcite{kwon2020pomo} to update the selected prompts in each batch. We use greedy inference, i.e., a deterministic trajectory $\tau$ is constructed iteratively based on the policy. In each construction step $t$, the next node $v_t$ is selected as the one with the maximum softmax probability $t = \arg\max_{i}(p_i)$ predicted by the decoder. $n$ trajectories are constructed from $n$ different starting points. 

The rewards $R(\tau_1),\dots,R(\tau_n)$ are the negative total distances of trajectories $\tau_1,\dots,\tau_n$. We use the following gradient ascent to update prompts $P$ in each batch with size $B$:
\begin{equation}
\begin{aligned}
    &\nabla_P J(\theta,P) \\ 
    &\approx \frac{1}{nB} \sum_{i=1}^B\sum_{j=1}^n\left(R\left(\tau^i_j \mid s \right)-b (s) \right) \nabla_P \log p_{\theta,P}\left(\tau^i_j \mid s \right),
\end{aligned}
\end{equation}
where $P$ and $\theta$ are trained prompts and fixed parameters for the model. $s$ represents the instances. $p_{\theta,P}(\tau^i_j)$ is the aggregation of the probability of selection in each step of the decoder based on both the fixed $\theta$ and the prompts $P$. $b(s)$ is the shared baseline~\cite{kwon2020pomo}.

\begin{table*}[tbp]
\renewcommand\arraystretch{1.2}
\centering
\resizebox{\textwidth}{!}{%
\begin{tabular}{llllllllllllll}
\toprule
& 50 CL & 50 EA & 50 EO & 50 IM & 50 GR & 50 MX & 100 CL& 100 EA& 100 EO& 100 IM& 100 GR& 100 MX& Avg.\\
   \midrule
HGS& 0   & 0   & 0   & 0   & 0   & 0   & 0   & 0   & 0   & 0   & 0   & 0   & 0   \\
LKH3 & 1.66\%& 1.53\%& 1.69\%& 1.81\%& 1.70\%& 1.62\%& 1.79\%& 1.61\%& 2.21\%& 3.61\%& 3.58\%& 3.80\%& 2.22\%\\
\midrule
Meta POMO     & 4.14\%& 3.56\%& 3.87\%& 3.95\%& 3.93\%& 3.58\%& 4.34\%& 3.73\%& 4.29\%& 4.32\%& 4.17\%& 3.80\%& 3.97\%\\
Omni & 4.61\%& 4.71\%& 5.20\%& 5.76\%& 5.63\%& 5.10\%& 3.32\%& 3.07\%& 3.81\%& 3.80\%& 3.67\%& 3.63\%& 4.36\%\\
Prompt     & \textbf{3.93\%} & \textbf{2.98\%} & \textbf{3.12\%} & \textbf{3.26\%} & \textbf{3.23\%} & \textbf{3.25\%} & \textbf{3.75\%} & \textbf{2.64\%} & \textbf{2.88\%} & \textbf{2.67\%} & \textbf{2.52\%} & \textbf{3.01\%} & \textbf{3.10\%} \\
\midrule
Meta POMO aug & 2.33\%& 2.05\%& 2.10\%& 2.24\%& 2.18\%& 2.00\%& 2.97\%& 2.48\%& 3.05\%& 3.07\%& 2.95\%& 2.60\%& 2.50\%\\
Omni aug       & 2.74\%& 2.80\%& 3.09\%& 3.51\%& 3.50\%& 2.96\%& 2.26\%& 1.98\%& 2.61\%& 2.66\%& 2.54\%& 2.46\%& 2.76\%\\
Prompt aug & \textbf{1.97\%} & \textbf{1.48\%} & \textbf{1.51\%} & \textbf{1.63\%} & \textbf{1.66\%} & \textbf{1.63\%} & \textbf{2.36\%} & \textbf{1.50\%} & \textbf{1.82\%} & \textbf{1.68\%} & \textbf{1.56\%} & \textbf{1.89\%} & \textbf{1.72\%} \\
\bottomrule
\end{tabular}% 
}
\caption{Zero-shot generalization performance on 12 new distributions.}~\label{table:zero-shot}
\end{table*}

\section{Experiments}

\subsection{Experimental Setting}

\paragraph{Model Settings}

We use the Attention model as our backbone pre-trained model. It is only trained on uniformly distributed CVRP instances of size 100. All the settings for the pre-trained model are the same as that used in the paper of Kwon~\shortcite{kwon2020pomo}. The number of encoder MHA layers is six and the decoder consists of one MHA and one SHA.

The settings of the key-prompt pair pool are as follows: The prompt size is set to be $M=16$, and the number of prompted tokens for each layer is $D=5$. As there are $L=6$ MHA layers in the encoder and the embedding size for each token is $E=128$, the lengths of the key and prompt vectors are $6 \cdot 128=768$ and $5 \cdot 6 \cdot 128=3840$, respectively.

\paragraph{Instance Generation}

We trained the model on a set of routing tasks with diverse sizes and geometrical distributions. The detailed instance generation is introduced in the Appendix, which is the same as that used by Zhou~\shortcite{zhou2023towards}. The problem size ranges from $50$ to $200$ with both uniform distribution and various clustered Gaussian distributions. There are in total $341$ distributions. 
%We want to not only show the fast adaptation ability of our prompt network but also demonstrate a good zero-shot generalization performance. 

\paragraph{Training Setup}

The prompts are trained with a batch size of $B=64$. The $341$ distributions are sequentially used during the training. In each batch, we randomly sample $B$ instanced from one distribution. As a result, every distribution will be sampled in $341$ epochs. The maximum number of epochs is $10,000$ and there are $1,000$ instances for each epoch. The learning rate decays from $1e-3$ to $1e-5$. It takes only about 2.5 days on a single RTX 2080Ti with 11 GB GPU memory.

\paragraph{Baselines}

We compared our proposed prompt learning to three different types of methods. 1) Baseline heuristic VRP solvers: hybrid genetic search (HGS)~\cite{vidal2013hybrid}, LKH3~\cite{helsgaun2017extension}; 2) NCO methods: basic POMO~\cite{kwon2020pomo} trained on single-distribution, and POMO~\cite{zhou2023towards} trained on diverse distribution; 3) meta learning NCO methods:  Meta-POMO~\cite{manchanda2023generalization}, and the newest revision of meta-learning method for omni-generalization on vehicle routing problems (Omni)~\cite{zhou2023towards}. The results for HGS and LKH are obtained by executing the open-source code on the instances. For the basic POMO, we train the model by ourselves on CVRP of size 100 with uniform distribution. This model is also utilized as the pre-trained model in our proposed prompt learning approach. For the meta-learning methods, we utilized the pre-trained model from Zhou~\shortcite{zhou2023towards} as it was trained on the same distributions as ours.

%\paragraph{Results on Training Tasks}

\subsection{Results on Training Tasks}

The results on training distributions are compared in Table~\ref{fig:train}. For simplicity, we list the averaged results over 1,000 instances on three problem sizes with uniform distribution. More results are included in the Appendix. We use HGS as the baseline and compare the training cost, average distances (Dis.), average gap to the baseline (Gap), and the inference time cost (Time). We executed HGS and LKH with time limits of 5s and 10s for problem sizes of 50 and 100, respectively. For instances with a problem size of 200, the time costs for each instance with HGS and LKH were 20s and 60s, respectively. It should be noted that LKH3 is not fully converged in some instances. We adopted the same data augmentation method (aug) as in Kwon~\shortcite{kwon2020pomo}. Additionally, for our prompt learning, we present the inference results using the top-k matched prompts. Further discussion on the top-k prompts is provided in the following section.

The proposed prompt learning method has a considerable reduction in training costs when compared to meta-learning methods. According to Zhou~\shortcite{zhou2023towards}, the second-order derivative method needs about 6 days and 71GB GPU memory and the training cost can be reduced to 3 days on 25GB GPU with a smarter strategy. For a fair comparison, we adjust the training cost of our prompt learning approach to match the experimental settings of the meta-learning methods. Specifically, prompt learning requires roughly 1 day on a 25GB GPU, considering the allowance for a larger batch size.

Our prompt learning model outperforms existing meta-learning methods and the basic POMO model on all three problem sizes in terms of average distances. The basic POMO model with single-distribution learning overfits the training distribution.  Because the basic POMO is trained on uniform distribution instances with a size of 100, it has good performance on the 100 instances set but deteriorates on the other two sizes. The two meta-learning methods' performance is more robust across different sizes compared with the basic POMO. Our prompt learning further reduces the gap. The prompt learning with the top 8 prompts ranks first in all sizes.

%\paragraph{Zero-shot Generalization}
\subsection{Zero-shot Generalization}

We demonstrate the zero-shot generalization performance of prompt learning on new distributions that were not used during training. We adopt the distribution proposed by Bi~\shortcite{bi2022learning}, which consists of a total of 12 different distributions, including cluster (CL), expansion (EA), explosion (EO), implosion (IM), grid (GR), and mixed (MX). Each distribution encompasses two different problem sizes, and we conduct tests on 1,000 instances for each distribution. We use HGS as the baseline. The total execution times on each distribution for both HGS and LKH on sizes 50 and 100 are 1.4h and 2.8h, respectively. 

Table~\ref{table:zero-shot} lists the zero-shot generalization performance on the 12 new distributions. The best results are in bold. Our prompt learning achieves the best average results. The average gap to the baseline is about 3\%. With data augmentation, the gap can be further reduced to less than 2\%.

%\paragraph{Top-k Prompts} 

\subsection{Top-k Prompts}

Instead of relying on a single best-matched prompt, we can employ multiple prompts simultaneously during the inference stage to enhance performance. To achieve this, we propose a top-k strategy, in which the top-k prompts (determined by the Euclidean distance between the key and feature vectors) are chosen. These k prompts are then used concurrently during inference, and the best solution is selected as the final solution for each instance. This approach allows us to fully leverage our learned prompts without incurring any additional training costs.

\begin{figure}[tbp]
    \vspace*{-10pt} % decrease the space below the figure
    \centering
    \includegraphics[width = 0.95\linewidth]{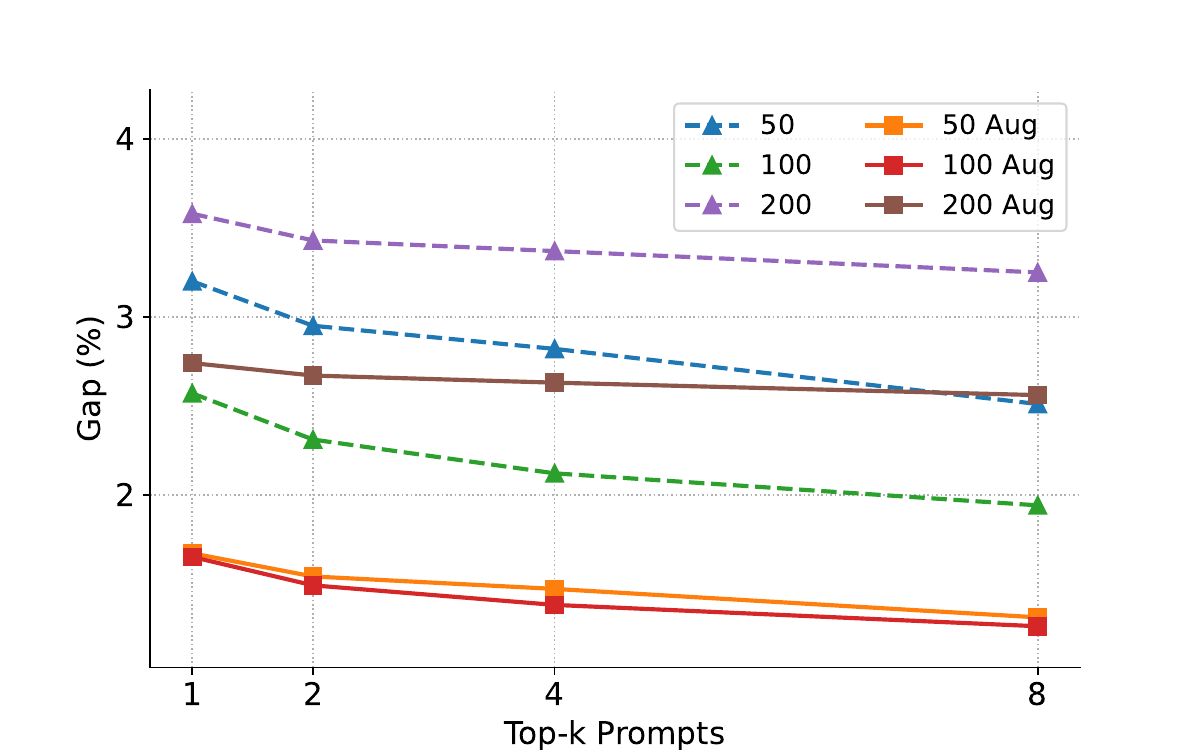}
    \caption{Results with different numbers of top-k prompts.}
    \label{fig:top_k}
\end{figure}

Figure~\ref{fig:top_k} shows the results obtained on instances of three different problem sizes with uniform distribution. The x-axis represents the number of prompts (k) employed in the top-k strategy, while the y-axis represents the difference in performance compared to the baseline HGS. Generally, the performance improves with an increase in the number of prompts. Moreover, the reduction in the gap is not linearly proportional to the number of prompts used, which suggests that the best-matched prompt is more significant than others.

% \begin{figure}[htbp]
%   \centering
%   \begin{minipage}{0.5\linewidth}  % Adjust the width of the figure as needed
%     \centering
%     \includegraphics[width = 1\linewidth]{Fig_topk.pdf}
%     \caption{Results with different numbers of top-k prompts.}
%     \label{fig:top_k}
%   \end{minipage}
%   \hfill
%   \begin{minipage}{0.48\linewidth}  % Adjust the width of the table as needed
%     \centering
%     \resizebox{0.9\linewidth}{!}{\begin{tabular}{|c|c|}
%       \hline
%       Column 1 & Column 2 \\
%       \hline
%       Row 1 & Row 1 \\
%       Row 2 & Row 2 \\
%       \hline
%     \end{tabular}
%     }
%     \caption{Table caption}
%     \label{tab:table_label}
%   \end{minipage}
% \end{figure}

% \begin{table}[t]%[htbp]
% \centering
% %\tiny
% \small
% \resizebox{0.30\textwidth}{!}{%
% \begin{tabular}{lccc}
% \toprule
% & 50 U  & 100 U & 200 U \\
% \midrule
% HGS       & 0      & 0      & 0      \\
% Top 1     & 3.20\% & 2.57\% & 3.58\% \\
% Top 1 aug & 1.67\% & 1.65\% & 2.74\% \\
% Top 2     & 2.95\% & 2.31\% & 3.43\% \\
% Top 2 aug & 1.54\% & 1.49\% & 2.67\% \\
% Top 4     & 2.82\% & 2.12\% & 3.37\% \\
% Top 4 aug & 1.47\% & 1.38\% & 2.63\% \\
% Top 8     & 2.51\% & 1.94\% & 3.25\% \\
% Top 8 aug & \textbf{1.31\%} & \textbf{1.26\% }& \textbf{2.56\%} \\
% \bottomrule
% \end{tabular}%
% }
% \caption{Results with different numbers of top-k prompts.}~\label{table:top_k}
% \end{table}

%\paragraph{Prompt Token Size} 

\subsection{Prompt Token Size} 

The number of prompt tokens $D$ in each encoder layer influences the performance of our prompt learning network. A larger number of tokens results in a longer prompt vector, providing the ability to prompt the attention-based encoder more effectively. In order to investigate the impact of the token number on our models, we conducted two additional prompt learning experiments. Specifically, we set the token numbers in the two models as 1 and 10, respectively. Consequently, the lengths of the prompt vectors in these models are $1 \cdot 6 \cdot 128=768$ and $10 \cdot 6 \cdot 128=7680$. All other training models and settings remain unchanged. 

The outcomes of the experiments involving different numbers of prompt tokens are presented in Table~\ref{table:tokensize}. The table includes results from four training distributions, distinguished by their numbers of prompt tokens. U and GM represent uniform distribution and Gaussian distribution with 3 clusters and scaled by 50 (details of instance generalization please refer to the Appendix). Minor differences in results are observed for instances with a size of 100, while more significant variations are noticed across other distributions. This discrepancy arises because the pre-trained basic POMO model utilized in prompt learning is trained on routing instances with a size of 100. Hence, the model already exhibits satisfactory performance on in-distribution instances. However, when adapting the pre-trained model to out-of-distribution instances, the number of tokens assumes importance. For example, in the case of 200 GM instances, there is approximately a 1\% performance increase (reduction in gap) from 1 token (2.93\%) to 10 tokens (2.02\%). Overall, prompt learning with a larger token size allows better generalization performance.

% A large number of tokens means a longer prompt vector and a better capability in prompting the attention-based encoder. We carried out two additional prompt learnings to study the influence of the number of tokens for each layer. The numbers of tokens for each layer in the two models are 1 and 10, respectively. As a result, the lengths of the prompt vector in the two models are $1*6*128=768$ and $10*6*128=7680$, respectively. We keep all the other models and training settings the same. Table~\ref{table:tokensize} lists the results of four training distributions with different numbers of prompt tokens. U and GM represent uniform distribution and Gaussian distribution with 3 clusters and scaled by 50 (details of instance generalization please refer to the Appendix). The results on instances with the size of 100 have minor differences, while the differences are more significant on other distributions. The reason is that the pre-trained basic POMO model in prompt learning is trained on routing instances of size 100. The model already performs well in in-distribution instances. When adapting the pre-trained model to out-of-distribution instances, the number of tokens is significant. For example, in 200 GM instances, the performance increase (gap reduction) reaches about 1\% from 1 token (2.93\%) to 10 tokens (2.02\%). In general, prompt learning with a large token size performs better than a small one. More prompt tokens enable prompting the pre-trained model to more diverse distributions.

\begin{table}[tbp]%[htbp]
\centering
\small
\resizebox{0.43\textwidth}{!}{%
\begin{tabular}{lcccc}
\toprule
& 50 U   & 100 U  & 200 U  & 200 GM\_50\_3 \\
\midrule
HGS       & 0      & 0      & 0      & 0 \\
Token 1     & 3.84\% & 2.46\% & 4.18\% & 4.45\%        \\
Token 1 aug & 2.00\% & 1.54\% & 3.27\% & 2.93\%        \\
Token 5     & 3.20\% & 2.57\% & 3.58\% & 3.30\%        \\
Token 5 aug & 1.67\% & 1.65\% & 2.74\% & 2.24\%        \\
Token 10     & 2.99\% & 2.59\% & 3.43\% & 2.97\%        \\
Token 10 aug & \textbf{1.47\% }& \textbf{1.67\%} & \textbf{2.60\%} & \textbf{2.02\% }     \\
\bottomrule
\end{tabular}%
}
\caption{Results with different prompt token sizes.}~\label{table:tokensize}
\end{table}

\subsection{Real-world Instances}

More experiments on real-world instances are conducted on five test suites: set A, B, P, X~\cite{uchoa2017new}, and XML~\cite{queiroga202110} from CVRPLIB~\footnote{http://vrp.atd-lab.inf.puc-rio.br/}. There are $115$ instances in total with various geometrical distributions, demands, and problem sizes, which can provide a comprehensive evaluation of our proposed method. Table~\ref{table:realworld} summarizes the average gap to the best-known results from CVRPLIB. The best results are in bold. The detailed results can be found in the Appendix.

\begin{table}[tbp]
\centering
\resizebox{0.48\textwidth}{!}{%
\begin{tabular}{cccccc}
\toprule
& POMO    & Meta POMO & Omni   & Prompt & Prompt top-8 \\
\midrule
A    & 7.3\%      & 2.3\%     & 4.4\%  & 2.1\%  & \textbf{1.8\%}     \\
B    & 12.6\%     & 1.9\%     & 2.4\%  & 1.7\%  & \textbf{1.5\%}     \\
P    & 35.6\%     & 12.9\%    & 10.8\% & 3.8\%  & \textbf{2.7\%}     \\
X   &	5.4\%	&4.9\%	&4.9\%	&4.7\%	& \textbf{3.5\%} \\
XML  & 4.4\%      & 5.4\%     & 5.8\%  & 6.1\%  & \textbf{3.4\%}     \\
\midrule
Average & 13.2\%     & 5.4\%     & 5.6\%  & 3.5\%  & \textbf{2.5\%}  \\
\bottomrule
\end{tabular}%
}
\caption{Results on CVRPLib Instances.}~\label{table:realworld}
\end{table}

\begin{figure}[tbp]
\vspace*{-5pt} % decrease the space below the figure
    \centering
    \includegraphics[width=0.45\textwidth]{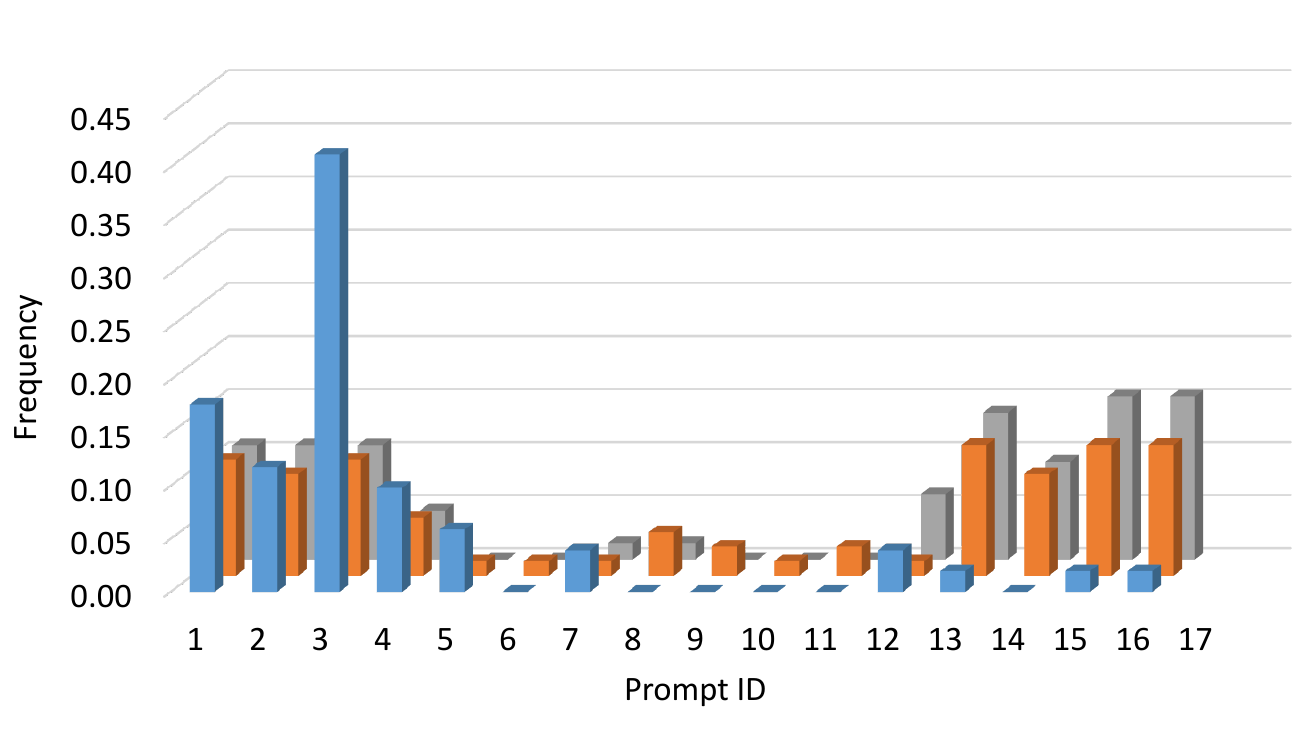}
    \caption{Selection frequencies of prompts on three different test sets. \textcolor{blue}{Blue:} Set P, \textcolor{orange}{Orange:} Set X, \textcolor{gray}{Grey:} Set XML.}
    \label{fig:prompt}
\end{figure}

In Figure~\ref{fig:prompt}, we visualize the selected frequencies (normalized) of 16 prompts on test set P, X, and XML. Set X and Set XML have similar frequency distributions, which are different from Set P. For example, prompts 11-15 are frequently used for Set X and XML while rarely used in Set P. The results show that the instances from Set X and Set XML are of similar distributions. As has been mentioned in the paper of Queiroga~\shortcite{queiroga202110}, the generator of XML is the same one used for X instances. It answers why the basic POMP performs well on these two sets while much worse on Set P. It also demonstrates that our prompt learning can recognize the features of new instances and select the best-matched prompt for better performance.

\section{Conclusion}

% %\paragraph{Conclusion} 
% \subsection{Conclusion}

This paper investigates the first prompt learning based neural combinatorial optimization (NCO) method to solve vehicle routing problems over diverse distributions. We propose a prompt-based attention network with a learnable key-prompt pair pool to facilitate the fast zero-shot adaptation of the pre-trained NCO model for cross-distribution generalization. To evaluate the effectiveness of our proposed prompt learning method, we conduct extensive experiments on test instances with various distributions. The results clearly demonstrate the superiority of our approach over classical single-distribution learning methods and existing meta learning techniques. Our prompt-based model achieves improvements in both in-distribution prediction and zero-shot generalization to a diverse set of new tasks while requiring lower training costs. 

\section*{Acknowledgments}

The work described in this paper was supported by the Research Grants Council of the Hong Kong Special Administrative Region, China (GRF Project No. CityU 11215723) and the Shenzhen Technology Plan, China (Grant No. JCYJ20220530113013031).

\bibliographystyle{named}
\bibliography{ijcai24}

\appendix
\section{Appendix}
\subsection{Model Details}
\subsubsection{Attention Mechanism}
The attention mechanism used in this work is proposed by Vaswani et al.~\shortcite{vaswani2017attention}. It involves mapping query ($Q$), key ($K$), and value ($V$) vectors to an output. For each node $i$, the query $Q_i$, key $K_i$, and value $V_i$ are projections of the input embedding $h_i$:

\begin{equation}
    Q_i = W^Q h_i, K_i = W^K h_i, V_i = W^V h_i,
\end{equation}
where $W^Q$, $W^K$, and $W^V$ are parameters of the respective sizes ($d_k \times d_h$) and ($d_v \times d_h$). The compatibility $u_{ij}$ is computed as:

\begin{equation}
    u_{ij} = \frac{Q_i^T K_j}{\sqrt{d_k}}.
\end{equation}
To obtain attention weights $a_{ij} \in [0,1]$, the compatibilities $u_{ij}$ are scaled using softmax:

\begin{equation}
    a_{ij} = \frac{e^{u_{ij}}}{\sum_j e^{u_{ij}}}.
\end{equation}
The output vector $h_i^o$ for node $i$ is a combination of the weights $a_{ij}$ and values $V_j$:

\begin{equation}
    h_i^o = \sum_j a_{ij} V_j.
\end{equation}

\subsubsection{Multi-head Attention (MHA)} The multi-head attention (MHA) mechanism allows the model to learn diverse information, leading to improved results. MHA consists of $h$ attention heads, each of which is an individual attention mechanism. The results from all heads are concatenated and then linearly projected:

\begin{equation}
\begin{aligned}
MHA (h_1, \dots, h_n) &= Concat (head_1, \dots, head_h) W^O \\
head_i &= Attention (h_1, \dots, h_n),
\end{aligned}
\end{equation}
where $W^O$ has size ($hd_v \times d_k$). In our experiments, we use 8 heads with different parameters, and the embedding size is set to 128. The parameter dimensions for each attention head in the model are $d_k = d_v = d_h/h = 16$.

\subsubsection{Encoder} The encoder consists of six layers of multi-head attention (MHA). As mentioned previously, the input to the encoder is the node features $f_i, i=1, \dots, n$. In our experiments, the input features for the $i$-th node are represented as $f_i = \{x_i, y_i, c_i\}$, where $(x_i, y_i)$ denote the coordinates and $c_i$ represents the demand. These input features are embedded through a linear projection to generate the initial feature embedding $h_i^{(0)}$. Each MHA layer in the encoder utilizes skip connections~\cite{he2016deep} and instance normalization (IN):
\begin{equation}
\begin{gathered}
\hat{h}_i^{(l)} = IN^l \left( h_i^{(l-1)} + MHA_i^l \left( h_1^{(l-1)}, \dots, h_n^{(l-1)} \right) \right), \\
h_i^{(l)} = IN^l \left( \hat{h}_i + F F^l \left( \hat{h}_i \right) \right),
\end{gathered}
\end{equation}
where $l$ and $l-1$ represent the current and previous embedding layers, respectively. The feedforward (FF) layer contains a hidden sublayer with ReLU activations. This encoding process generates the final node embeddings $h_i^{(L)}$. It is performed only once, and the static node embeddings are reused for every decoding step.

\begin{figure*}[t]
    \centering
    \includegraphics[width=0.9\textwidth]{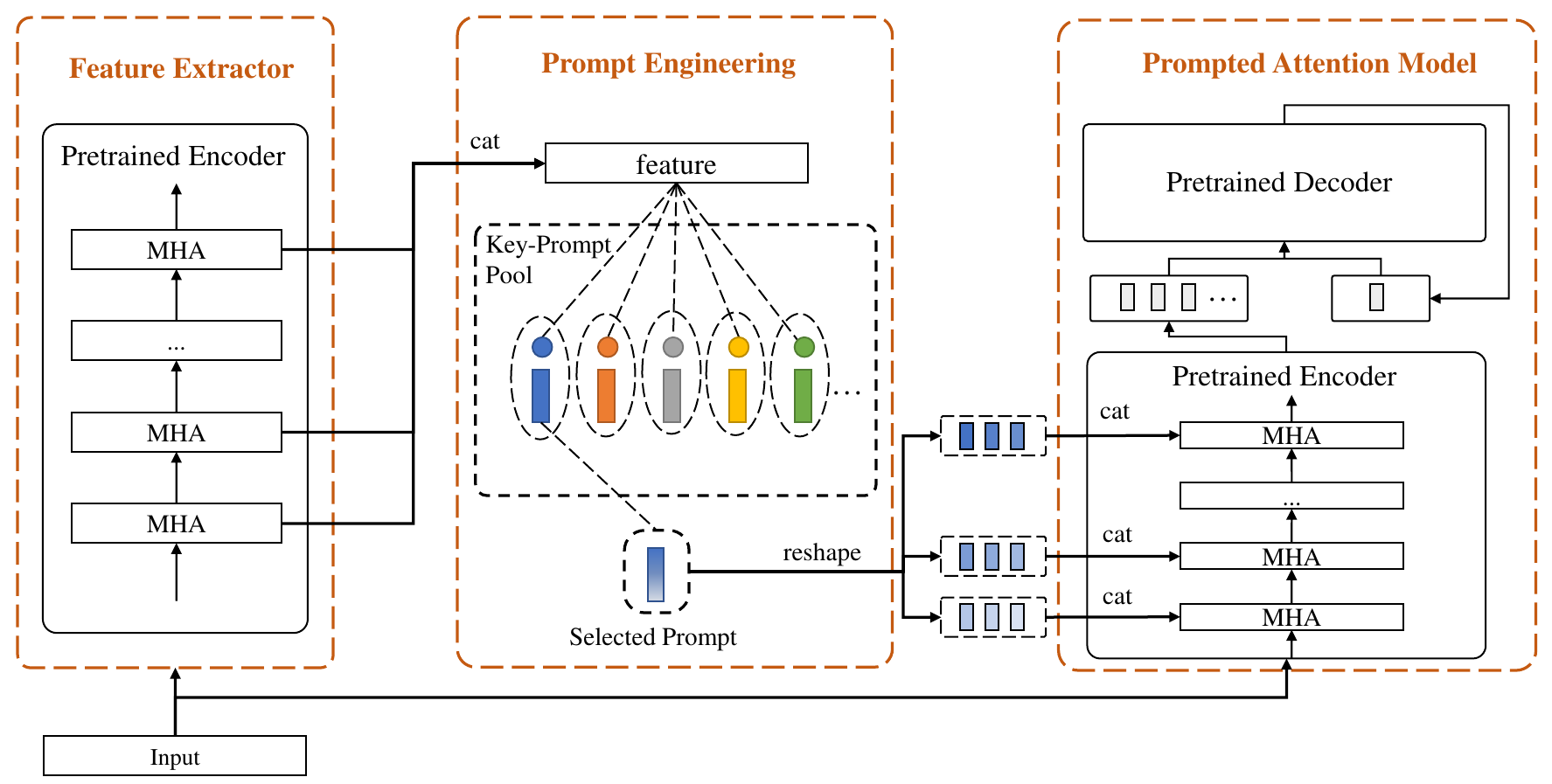}
    \caption{Model structure of our proposed prompt learning method, which consists of three main parts. \textbf{1) Feature Extractor:} We use a pre-trained encoder to extract the feature for a given input instance, which is defined as the concatenation of multiple MHA outputs for different layers. \textbf{2) Prompt Engineering}: The most suitable key is selected to match the extracted feature of the input instance, and then its associated prompt will be used to adjust the pre-trained NCO model in a zero-shot manner. \textbf{3) Prompted Neural Solver:} The prompt embedding is decomposed into $L$ subprompts, of which each one consists of $D$ tokens. Each subprompt will be concatenated into each corresponding layer in the pre-trained encoder. In this way, the pre-trained NCO model is fast adjusted to better tackle the input problem instance.}
    \label{fig:model_structure}
\end{figure*}

\subsubsection{Decoder} The decoder in this work consists of a multi-head attention (MHA) layer followed by a self-attention (SHA) layer, as proposed by Kool et al.~\shortcite{kool2018attention}. The computation of queries, keys, and values for the MHA layer is as follows:
\begin{equation}
\begin{aligned}
Q_c &= W^Q h_c, K_i = W^K h_i, V_i = W^V h_i, \\
h_c &= Concat(h_t, a_t),
\end{aligned}
\end{equation}
where $h_t$ is the embedding of the current visited node and $a_t$ is the attribute vector. $h_i$ represents the output embedding from the encoder for node $i$.

In the SHA layer, the compatibility $u_{cj}$ is computed using equation (5), and the results are clipped within the range [-10,10] using the tanh function. Compatibility values for masked nodes are set to $-\infty$ to exclude them:
\begin{equation}
u_{cj} = \left\{
\begin{aligned}
& 10 \cdot \tanh \left( \frac{q_c^T k_j}{\sqrt{d_k}} \right) \quad &\text{if } j \notin m_t \\
& -\infty \quad &\text{otherwise}.
\end{aligned}
\right.
\end{equation}
The output probability of selecting the next node is computed as the softmax of the output compatibilities $p_i = \frac{e^{u_i}}{\sum_j e^{u_j}}$.

\subsection{Prompt Learning Illustration}~\label{sec:prompt}

A more comprehensive illustration of the prompt learning framework is presented in Figure~\ref{fig:model_structure}. As previously described in the main paper, the framework comprises three components: 1) feature extractor, 2) prompt engineering, and 3) prompted neural solver. Please refer to the method section in the main paper for detailed explanations of each component.

\subsection{Instance Generation and Utilization}~\label{sec:instances}

\paragraph{Gaussian Mixture Distribution} 

% We use the same Gaussian mixture distribution as~\cite{zhou2023towards}. It is denoted as $GM_c^l$, where $c$ and $l$ represent the cluster number and the scale, respectively. The instances are generated in four steps: 1) the coordinate of the center of each cluster is uniformly sampled from $U(0,l)$. 2) other nodes are equally distributed into $c$ clusters. 3) for the nodes in each cluster the coordinates are generated from a Gaussian distribution. 4) The range of coordinates is scaled into $[0,1]$ using min-max normalization.

We use the same Gaussian mixture distribution to generate instances as in Zhou et al. \shortcite{zhou2023towards}. The distribution is denoted as $GM_c^l$, where $c$ and $l$ represent the cluster number and the scale, respectively. The instances are generated in four steps: 

\begin{enumerate}
    \item Uniformly sample the coordinate of the center of each cluster from $U(0,l)$. 
    
    \item Evenly distribute other nodes into $c$ clusters. 
    
    \item Generate the coordinates for the nodes from a separate Gaussian distribution for each cluster.
    
    \item Scale the range of coordinates for all nodes into $[0,1]$ using min-max normalization. 
\end{enumerate}

\paragraph{Training Distributions} We use the same training distributions as in Zhou et al.~\shortcite{zhou2023towards}. The problem sizes that we use are $\{50,55,60,\dots,195,200\}$. For each problem size, we have $11$ different geometrical distributions $d(c, l) \in \mathcal{D}=\{(0,0),(1,1)\} \cup\{c[3,5,7] \times l[10,30,50]\}$, where $(0,0)$ represents uniform distribution and $(1,1)$ represents single Gaussian distribution. The rest $9$ distributions are Gaussian mixture distributions with difference combinations of $c$ and $l$. The settings of demand follow Kool et al.~\shortcite {kool2018attention}, where the demand of each node is randomly sampled from integers $\{1,2,\dots,9\}$. For the distribution with problem size n, the vehicle capacity is set to be $C= \lceil 30+ \frac{n}{5} \rceil$. The demands are normalized by the capacity. 

\paragraph{Sequential Training} 

% In training, we use one distribution for each epoch. The $341$ distributions are sequentially used.The order is firstly determined by the problem size and then the distribution set $\mathcal{D}=\{(0,0),(1,1),(3,10),\dots,(7,50) \}$. As a result, the uniform distribution $(0,0)$ with problem size 50 will be used in the first epoch, and the Gaussian mixture distribution $GM_10^50$ with problem size 200 will be used in the 341 epoch. 

We sequentially use one distribution from all $341$ distributions at each epoch for training our model. The order is firstly determined by the problem size and then the distribution set $\mathcal{D}=\{(0,0),(1,1),(3,10),\dots,(7,50) \}$. As a result, the uniform distribution $(0,0)$ with problem size $50$ will be used in the first epoch, and the Gaussian mixture distribution $GM_{10}^{50}$ with problem size $200$ will be used in the $341$-th epoch. 

\paragraph{Testing Instance Generation}

%There are in total 12 distributions (combinations of 6 geometrical distributions and 2 problem sizes) used in the testing of zero-shot generalization. The geomatical distributions include cluster, expansion, explosion, implosion, grid, and mixed. We use the dataset provided in~\cite{bi2022learning}. The detailed formulations can be found in~\cite{bossek2019evolving}.

\begin{figure}[tbp]
    \centering
    \includegraphics[width=0.47\textwidth]{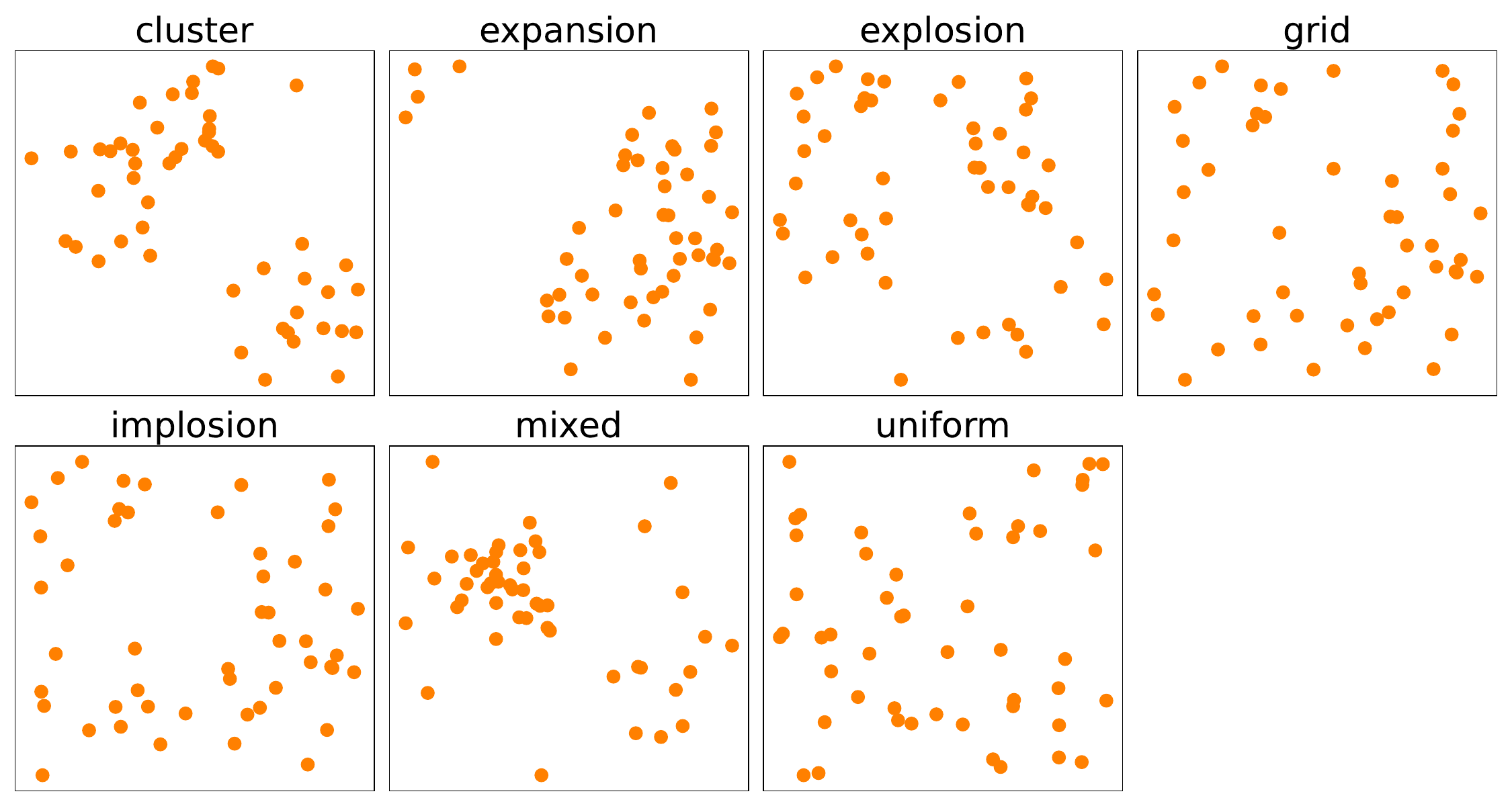}
    \caption{Illustration of six geometrical distributions used in testing and the uniform distribution.}
    \label{fig:testing}
\end{figure}

We use 12 different distributions (combinations of 6 geometrical distributions and 2 problem sizes) in total to test the zero-shot generalization performance for all methods. As is illustrated in Figure~\ref{fig:testing}, the geometrical distributions include cluster, expansion, explosion, implosion, grid, and mixed. We use the dataset provided in Bi et al.~\shortcite{bi2022learning} with detailed formulations from Bossek et al.~\shortcite{bossek2019evolving}.

% \section{Key-Prompt Initialization}

% \paragraph{Feature Data Generation} 

% We randomly sample 128 instances for each distribution, which results in $43,648$ instances to generate the feature data. For each instance $i$, we use the same feature extractor as introduced in the main paper to extract features $F_i$:
% \begin{equation}
%     \begin{gathered}
%         F_l = \frac{1}{n} \sum_{i=1}^n {\left( \hat{h}_1^{(l)} + MHA^l\left(\hat{h}_i\right) \right)} \\
%         F = cat\{ F_1 ,F_2, \dots,F_L  \}
%     \end{gathered}
% \end{equation}

% The normalized features will be used as the samples for clustering. Since each feature is a vector of length $768$, the size of the training dataset is $768*43,648$.

% \paragraph{Generation of Keys} 

% We use the cluster centers of the features as the keys. We first divide the $43,648$ samples into four groups with respect to the problem sizes. For each group, K-means is used to cluster the $10,907$ samples into $N$ clusters. In the end, we will have $M = 4*N$ cluster centers. These cluster centers are vectors of length $768$, which will be used as the keys in prompt learning.

% \paragraph{Initialization of Prompts}

% For each key $K_i$, we randomly generate a prompt $P_i$ and scale the prompt into the range $[-1,1]$. The keys $K_i,i=1,\dots,M$ are fixed during the whole training process, and we only learn the prompts $P_i,i=1,\dots,M$.

% % \section{Prompted encoder}

\begin{figure}[tbp]
    \centering
    \includegraphics[width=0.48\textwidth]{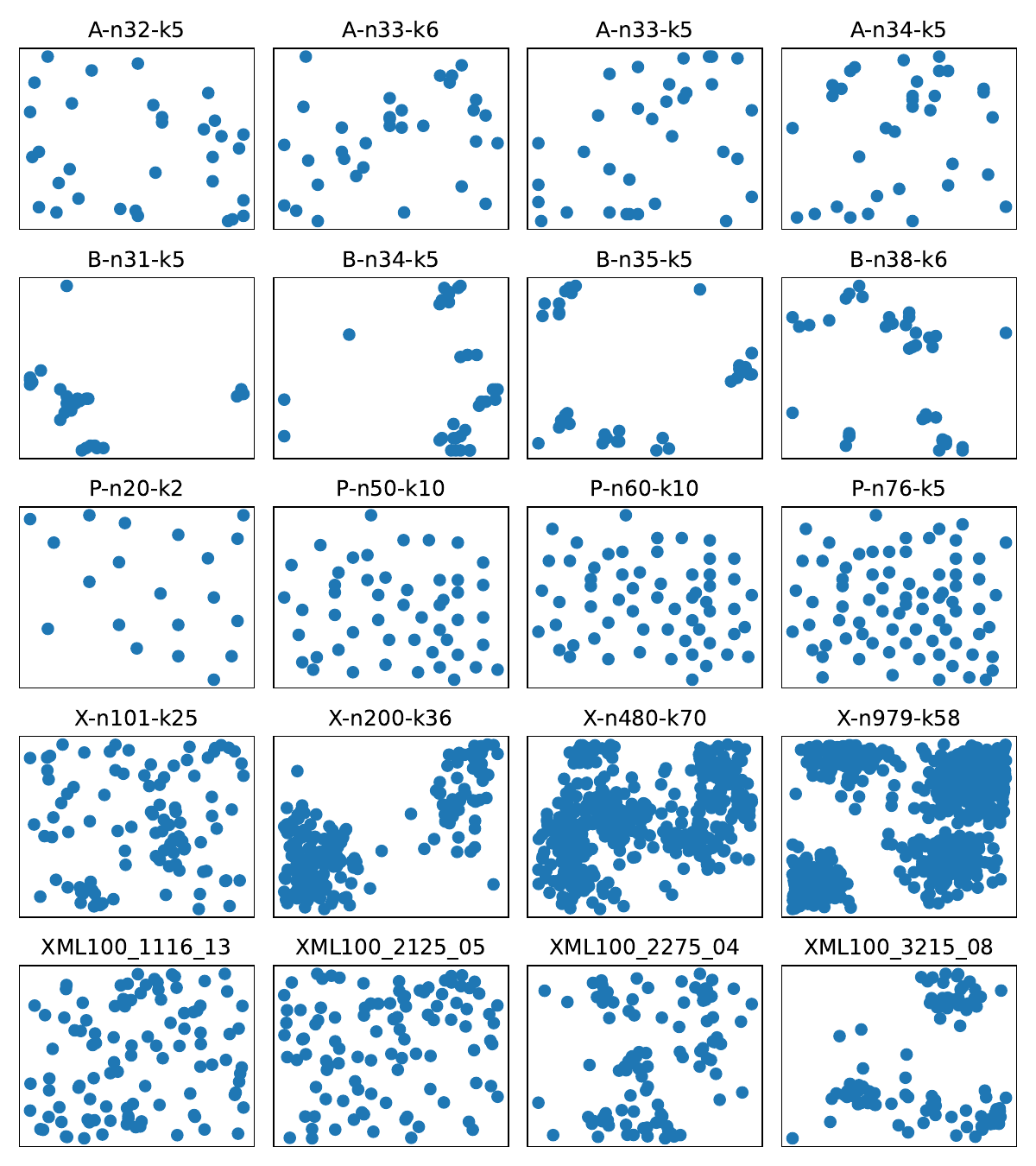}
    \caption{Illustration of CVRPLIB sets with diverse distributions and sizes. For the set A, B, P, and X, $n$ and $k$ represent the number of nodes and the minimum number of vehicles, respectively. The notations for XML instances represent different distributions.}
    \label{fig:cvrplib}
\end{figure}

\subsection{Additional Results and Discussions}~\label{sec:cvrplib}

We conduct extensive additional experiments to compare our proposed prompt learning with existing neural constructive NCO methods for vehicle routing problems. 

\begin{table*}[htbp]
\centering
\resizebox{0.8\textwidth}{!}{%
\begin{tabular}{lccccccccccccc}
\toprule
\multicolumn{1}{c}{Instance} & Baseline & \multicolumn{2}{c}{POMO} & \multicolumn{2}{c}{POMO Multi}       & \multicolumn{2}{c}{Meta POMO}  & \multicolumn{2}{c}{Omni}       & \multicolumn{2}{c}{Prompt (Ours)} & \multicolumn{2}{c}{Prompt top-8 (Ours)} \\
\midrule
A-n32-k5    & 784      & 895 & 14.2\%         & 817 & 4.2\%& 831 & 6.0\%& 853 & 8.8\%& \textbf{802}    & \textbf{2.3\%}  & \textbf{802}     & \textbf{2.3\%}   \\
A-n33-k5    & 661      & 828 & 25.3\%         & \textbf{672}  & \textbf{1.7\%} & \textbf{672}  & \textbf{1.7\%} & 692 & 4.7\%& 676   & 2.3\% & 676    & 2.3\%  \\
A-n33-k6    & 742      & 937 & 26.3\%         & \textbf{747}  & \textbf{0.7\%} & 751 & 1.2\%& 784 & 5.7\%& 755   & 1.8\% & 750    & 1.1\%  \\
A-n34-k5    & 778      & 952 & 22.4\%         & \textbf{788}  & \textbf{1.3\%} & 790 & 1.5\%& 809 & 4.0\%& 790   & 1.5\% & 790    & 1.5\%  \\
A-n36-k5    & 799      & 890 & 11.4\%         & 815 & 2.0\%& \textbf{806}  & \textbf{0.9\%} & 872 & 9.1\%& 815   & 2.0\% & 815    & 2.0\%  \\
A-n37-k5    & 669      & 741 & 10.8\%         & 698 & 4.3\%& 705 & 5.4\%& 719 & 7.5\%& \textbf{695}    & \textbf{3.9\%}  & \textbf{695}     & \textbf{3.9\%}   \\
A-n37-k6    & 949      & 1015& 6.9\%& 985 & 3.8\%& 975 & 2.7\%& 981 & 3.4\%& \textbf{972}    & \textbf{2.4\%}  & \textbf{972}     & \textbf{2.4\%}   \\
A-n38-k5    & 730      & 808 & 10.6\%         & 766 & 4.9\%& \textbf{745}  & \textbf{2.1\%} & 767 & 5.1\%& 748   & 2.5\% & 748    & 2.5\%  \\
A-n39-k5    & 822      & 910 & 10.7\%         & 843 & 2.6\%& \textbf{830}  & \textbf{1.0\%} & 863 & 5.0\%& 835   & 1.6\% & 835    & 1.6\%  \\
A-n39-k6    & 831      & 895 & 7.6\%& 839 & 1.0\%& 847 & 1.9\%& 886 & 6.6\%& 844   & 1.6\% & \textbf{838}     & \textbf{0.8\%}   \\
A-n44-k6    & 937      & 992 & 5.8\%& \textbf{952}  & \textbf{1.6\%} & 962 & 2.7\%& 972 & 3.7\%& 958   & 2.2\% & 954    & 1.8\%  \\
A-n45-k6    & 944      & 974 & 3.2\%& \textbf{961}  & \textbf{1.8\%} & 963 & 2.0\%& 986 & 4.4\%& 970   & 2.8\% & 967    & 2.4\%  \\
A-n45-k7    & 1146     & 1183& 3.2\%& 1167& 1.8\%& 1161& 1.3\%& 1189& 3.8\%& 1163  & 1.5\% & \textbf{1154}    & \textbf{0.7\%}   \\
A-n46-k7    & 914      & 934 & 2.2\%& 932 & 2.0\%& 928 & 1.5\%& 937 & 2.5\%& \textbf{920}    & \textbf{0.7\%}  & \textbf{920}     & \textbf{0.7\%}   \\
A-n48-k7    & 1073     & 1126& 4.9\%& 1130& 5.3\%& 1108& 3.3\%& 1114& 3.8\%& 1104  & 2.9\% & \textbf{1101}    & \textbf{2.6\%}   \\
A-n53-k7    & 1010     & 1056& 4.6\%& 1066& 5.5\%& 1058& 4.8\%& 1065& 5.4\%& \textbf{1039}   & \textbf{2.9\%}  & \textbf{1039}    & \textbf{2.9\%}   \\
A-n54-k7    & 1167     & 1201& 2.9\%& 1208& 3.5\%& \textbf{1176} & \textbf{0.8\%} & 1224& 4.9\%& 1181  & 1.2\% & 1181   & 1.2\%  \\
A-n55-k9    & 1073     & 1120& 4.3\%& 1101& 2.6\%& \textbf{1087} & \textbf{1.3\%} & 1099& 2.4\%& 1092  & 1.8\% & 1088   & 1.4\%  \\
A-n60-k9    & 1354     & 1370& 1.2\%& 1379& 1.8\%& 1374& 1.5\%& 1387& 2.4\%& 1376  & 1.6\% & \textbf{1368}    & \textbf{1.0\%}   \\
A-n61-k9    & 1034     & 1077& 4.1\%& 1080& 4.4\%& 1066& 3.1\%& 1069& 3.4\%& 1066  & 3.1\% & \textbf{1053}    & \textbf{1.8\%}   \\
A-n62-k8    & 1288     & \textbf{1310} & \textbf{1.7\%} & 1326& 3.0\%& 1319& 2.4\%& 1350& 4.8\%& 1315  & 2.1\% & 1312   & 1.9\%  \\
A-n63-k9    & 1616     & 1651& 2.2\%& 1650& 2.1\%& \textbf{1640} & \textbf{1.5\%} & 1652& 2.2\%& 1648  & 2.0\% & 1642   & 1.6\%  \\
A-n63-k10   & 1314     & 1329& 1.1\%& 1350& 2.7\%& 1343& 2.2\%& 1367& 4.0\%& 1333  & 1.4\% & \textbf{1328}    & \textbf{1.1\%}   \\
A-n64-k9    & 1401     & 1433& 2.3\%& 1441& 2.9\%& 1443& 3.0\%& 1459& 4.1\%& \textbf{1432}   & \textbf{2.2\%}  & \textbf{1432}    & \textbf{2.2\%}   \\
A-n65-k9    & 1174     & 1213& 3.3\%& 1213& 3.3\%& \textbf{1199} & \textbf{2.1\%} & \textbf{1199} & \textbf{2.1\%} & 1205  & 2.6\% & 1204   & 2.6\%  \\
A-n69-k9    & 1159     & 1188& 2.5\%& 1197& 3.3\%& 1185& 2.2\%& 1205& 4.0\%& 1187  & 2.4\% & \textbf{1176}    & \textbf{1.5\%}   \\
A-n80-k10   & 1763     & 1801& 2.2\%& \textbf{1789} & \textbf{1.5\%} & 1802& 2.2\%& 1790& 1.5\%& 1792  & 1.6\% & \textbf{1789}    & \textbf{1.5\%}   \\
\midrule
Average    &&     & 7.3\%&     & 2.8\%&     & 2.3\%&     & 4.4\%&& 2.1\% & & \textbf{1.8\%}  \\
\bottomrule
\end{tabular}%
}
\caption{Results on CVRPLIB Set A instances.}~\label{table:setA}
\end{table*}

\paragraph{Compared Methods}
The compared methods are as follows:
\begin{itemize}
    \item \textbf{POMO}~\cite{kwon2020pomo}: POMO trained on single distribution (uniform distribution with problem size 100).
    \item \textbf{POMO Multi}~\cite{kwon2020pomo}: POMO trained on the same training distributions as ours.
    \item \textbf{Meta POMO}~\cite{manchanda2023generalization}: POMO trained with the meta-learning method proposed in~\cite{manchanda2023generalization} on the same training distributions as ours.
    \item \textbf{Omni}~\cite{zhou2023towards}: POMO trained with the meta-learning method proposed in Zhou et al.~\shortcite{zhou2023towards} on the same training distributions as ours.
    \item \textbf{Prompt}: our proposed prompt learning.
    \item \textbf{Prompt top-8} : our proposed prompt learning with top-8 prompts.
\end{itemize}

\paragraph{Test Sets}

The experiments are conducted on five test suites: Sets A, B, P, X~\cite{uchoa2017new}, and XML~\cite{queiroga202110}. Most of these sets were extracted from real-world problems, and XML was proposed recently for testing learning methods in vehicle routing. In total, there are 115 instances with various geometrical distributions, demands, and problem sizes, which allow for a comprehensive evaluation of our proposed method.

All instance data were obtained from CVRPLIB~\footnote{http://vrp.atd-lab.inf.puc-rio.br/}. The baseline results for Sets A, B, P, and X correspond to the best-known results from CVRPLIB. The best-known results were achieved by first minimizing the number of vehicles and then minimizing the total distance. In some instances, the best-known results do not necessarily represent the solution with the shortest total distance, as they may utilize more vehicles. Consequently, in a few cases (e.g., B-n51-k7 and P-n55-k15), we obtained better results than the best-known solution (baseline). The baseline results for XML instances were provided by the original paper~\cite{queiroga202110}, which employed a state-of-the-art branch-cut-and-price algorithm~\cite{pessoa2020generic}.

For each instance, the customer coordinates were normalized within the unit range of [0,1]. The demands were also normalized with respect to the vehicle capacity.

\paragraph{Distribution}

Figure~\ref{fig:cvrplib} illustrates the different instance sets of CVRPLIB, which possess diverse distributions and sizes. For example, the nodes in Set B are clustered together, whereas the nodes in Set P are more sparsely distributed. Sets X and XML exhibit similar patterns. As discussed in Figure 3 of the main paper, our prompt learning approach is capable of recognizing the features of new instances and selecting the best-matched prompt to achieve superior performance.

\paragraph{Results and Discussion}

Table~\ref{table:setA} to Table~\ref{table:setXML} present the results, comparing the distances of the solutions generated by different NCO methods and the gap between these distances and the baselines. The best results are indicated in bold.

Overall, our prompt learning approach outperforms all the methods compared in the study. With the top-8 prompts, the average gap is less than 2\% for Sets A and B.  POMO performs well in X instances and XML instances, but is significantly worse in the other three test sets. This is because it was only trained on a uniform distribution with a problem size of 100. The two meta-learning methods exhibit robust performance across various distributions; however, their performance is inferior to our proposed prompt learning approach. On test Set P, the average gap of our prompt learning approach is approximately 3\%, whereas the gaps of the meta-learning methods exceed 10\%. 

\begin{table*}[htbp]
\centering

\resizebox{0.9\textwidth}{!}{%
\begin{tabular}{lccccccccccccc}
\toprule
\multicolumn{1}{c}{Instance} & Baseline & \multicolumn{2}{c}{POMO} & \multicolumn{2}{c}{POMO Multi}       & \multicolumn{2}{c}{Meta POMO}   & \multicolumn{2}{c}{Omni}       & \multicolumn{2}{c}{Prompt (Ours)} & \multicolumn{2}{c}{Prompt top-8 (Ours)} \\
\midrule
B-n31-k5    & 672      & 992 & 47.6\%         & 685 & 1.9\%& \textbf{673}  & \textbf{0.1\%}  & \textbf{673}  & \textbf{0.1\%} & 685   & 1.9\% & 682   & 1.5\%   \\
B-n34-k5    & 788      & 1057& 34.2\%         & 794 & 0.8\%& 797 & 1.1\% & 799 & 1.4\%& 793   & 0.6\% & \textbf{792}    & \textbf{0.5\%}    \\
B-n35-k5    & 955      & 1257& 31.6\%         & 977 & 2.3\%& \textbf{964}  & \textbf{0.9\%}  & 999 & 4.6\%& 976   & 2.2\% & 970   & 1.6\%   \\
B-n38-k6    & 805      & 1087& 35.0\%         & 822 & 2.1\%& 814 & 1.1\% & \textbf{810}  & \textbf{0.6\%} & 820   & 1.9\% & 814   & 1.1\%   \\
B-n39-k5    & 549      & 786 & 43.1\%         & \textbf{552}  & \textbf{0.5\%} & 553 & 0.7\% & 557 & 1.5\%& 554   & 0.9\% & 554   & 0.9\%   \\
B-n41-k6    & 829      & 911 & 9.9\%& 849 & 2.4\%& 843 & 1.7\% & 838 & 1.1\%& \textbf{837}    & \textbf{1.0\%}  & \textbf{837}    & \textbf{1.0\%}    \\
B-n43-k6    & 742      & 798 & 7.6\%& 755 & 1.8\%& 757 & 2.0\% & 765 & 3.1\%& \textbf{749}    & \textbf{0.9\%}  & \textbf{749}    & \textbf{0.9\%}    \\
B-n44-k7    & 909      & 939 & 3.3\%& 934 & 2.8\%& 944 & 3.9\% & 932 & 2.5\%& \textbf{925}    & \textbf{1.8\%}  & \textbf{924}    & \textbf{1.7\%}    \\
B-n45-k5    & 751      & 770 & 2.5\%& \textbf{757}  & \textbf{0.8\%} & 763 & 1.6\% & 773 & 2.9\%& 758   & 0.9\% & 758   & 0.9\%   \\
B-n45-k6    & 678      & 749 & 10.5\%         & 720 & 6.2\%& 719 & 6.0\% & 718 & 5.9\%& \textbf{689}    & \textbf{1.6\%}  & \textbf{689}    & \textbf{1.6\%}    \\
B-n50-k7    & 741      & 789 & 6.4\%& 753 & 1.6\%& \textbf{746}  & \textbf{0.7\%}  & 751 & 1.3\%& 751   & 1.3\% & 748   & 0.9\%   \\
B-n50-k8    & 1312     & 1346& 2.6\%& 1336& 1.8\%& 1336& 1.8\% & 1335& 1.8\%& \textbf{1331}   & \textbf{1.4\%}  & \textbf{1331}   & \textbf{1.4\%}    \\
B-n51-k7    & 1032     & 1200& 16.3\%         & 1026& -0.6\%         & \textbf{1020} & \textbf{-1.2\%} & 1024& -0.8\%         & 1024  & -0.8\%& 1021  & -1.1\%  \\
B-n52-k7    & 747      & 767 & 2.7\%& 756 & 1.2\%& \textbf{752}  & \textbf{0.7\%}  & 760 & 1.7\%& 757   & 1.3\% & 756   & 1.2\%   \\
B-n56-k7    & 707      & 743 & 5.1\%& 726 & 2.7\%& \textbf{724}  & \textbf{2.4\%}  & 732 & 3.5\%& 729   & 3.1\% & \textbf{724}    & \textbf{2.4\%}    \\
B-n57-k7    & 1153     & 1160& 0.6\%& 1163& 0.9\%& 1154& 0.1\% & 1154& 0.1\%& 1158  & 0.4\% & \textbf{1149}   & \textbf{-0.3\%}   \\
B-n57-k9    & 1598     & 1637& 2.4\%& 1634& 2.3\%& 1612& 0.9\% & 1639& 2.6\%& \textbf{1609}   & \textbf{0.7\%}  & \textbf{1609}   & \textbf{0.7\%}    \\
B-n63-k10   & 1496     & 1582& 5.7\%& 1534& 2.5\%& 1523& 1.8\% & 1561& 4.3\%& \textbf{1513}   & \textbf{1.1\%}  & \textbf{1513}   & \textbf{1.1\%}    \\
B-n64-k9    & 861      & 923 & 7.2\%& \textbf{898}  & \textbf{4.3\%} & 900 & 4.5\% & 908 & 5.5\%& 909   & 5.6\% & 905   & 5.1\%   \\
B-n66-k9    & 1316     & 1339& 1.7\%& \textbf{1328} & \textbf{0.9\%} & 1336& 1.5\% & 1335& 1.4\%& 1331  & 1.1\% & 1331  & 1.1\%   \\
B-n67-k10   & 1032     & 1111& 7.6\%& 1091& 5.7\%& 1069& 3.6\% & 1095& 6.1\%& 1065  & 3.2\% & \textbf{1063}   & \textbf{3.0\%}    \\
B-n68-k9    & 1272     & 1303& 2.4\%& \textbf{1296} & \textbf{1.9\%} & 1310& 3.0\% & 1299& 2.1\%& 1305  & 2.6\% & 1305  & 2.6\%   \\
B-n78-k10   & 1221     & 1268& 3.8\%& 1247& 2.1\%& 1267& 3.8\% & \textbf{1246} & \textbf{2.0\%} & 1271  & 4.1\% & 1263  & 3.4\%   \\
\midrule
Average     &&     & 12.6\%         &     & 2.1\%&     & 1.9\% &     & 2.4\%&& 1.7\% && \textbf{1.5\%}   \\
\bottomrule
\end{tabular}%
}
\caption{Results on CVRPLIB Set B instances.}~\label{table:setB}
\end{table*}

\begin{table*}[htbp]
\centering

\resizebox{0.9\textwidth}{!}{%
\begin{tabular}{lccccccccccccc}
\toprule
\multicolumn{1}{c}{Instance} & Baseline & \multicolumn{2}{c}{POMO} & \multicolumn{2}{c}{POMO Multi}      & \multicolumn{2}{c}{Meta POMO} & \multicolumn{2}{c}{Omni} & \multicolumn{2}{c}{Prompt (Ours)} & \multicolumn{2}{c}{Prompt top-8 (Ours)} \\
\midrule
P-n16-k8 & 450      & 490& 8.9\% & 453& 0.7\%& 496& 10.2\%         & 503        & 11.8\%      & \textbf{452}   & \textbf{0.4\%}   & \textbf{452}    & \textbf{0.4\%}    \\
P-n19-k2 & 212      & 476& 124.7\%         & \textbf{224} & \textbf{5.7\%} & 281& 32.5\%         & 250        & 17.9\%      & 228  & 7.5\%  & 228   & 7.5\%   \\
P-n20-k2 & 216      & 562& 160.3\%         & 233& 7.9\%& 295& 36.6\%         & 300        & 38.9\%      & 230  & 6.5\%  & \textbf{228}    & \textbf{5.6\%}    \\
P-n21-k2 & 211      & 592& 180.8\%         & 228& 8.1\%& 316& 49.8\%         & 309        & 46.4\%      & \textbf{219}   & \textbf{3.8\%}   & \textbf{219}    & \textbf{3.8\%}    \\
P-n22-k2 & 216      & 519& 140.2\%         & 239& 10.6\%         & 355& 64.4\%         & 274        & 26.9\%      & \textbf{219}   & \textbf{1.4\%}   & \textbf{219}    & \textbf{1.4\%}    \\
P-n22-k8 & 603      & 799& 32.5\%& 746& 23.7\%         & 786& 30.3\%         & 731        & 21.2\%      & 634  & 5.1\%  & \textbf{591}    & \textbf{-2.0\%}   \\
P-n23-k8 & 529      & 731& 38.2\%& 541& 2.3\%& 642& 21.4\%         & 587        & 11.0\%      & \textbf{539}   & \textbf{1.9\%}   & \textbf{539}    & \textbf{1.9\%}    \\
P-n40-k5 & 458      & 577& 26.0\%& \textbf{460} & \textbf{0.4\%} & 470& 2.6\%& 491        & 7.2\%       & 473  & 3.3\%  & 473   & 3.3\%   \\
P-n45-k5 & 510      & 591& 15.9\%& \textbf{517} & \textbf{1.4\%} & 525& 2.9\%& 534        & 4.7\%       & 524  & 2.7\%  & 524   & 2.7\%   \\
P-n50-k7 & 554      & 614& 10.8\%& 573& 3.4\%& 570& 2.9\%& 587        & 6.0\%       & 571  & 3.1\%  & \textbf{569}    & \textbf{2.7\%}    \\
P-n50-k8 & 631      & 676& 7.2\% & 641& 1.6\%& 642& 1.7\%& 644        & 2.1\%       & 637  & 1.0\%  & \textbf{636}    & \textbf{0.8\%}    \\
P-n50-k10& 696      & 752& 8.0\% & 716& 2.9\%& 715& 2.7\%& 723        & 3.9\%       & \textbf{711}   & \textbf{2.2\%}   & \textbf{711}    & \textbf{2.2\%}    \\
P-n51-k10& 741      & 771& 4.1\% & \textbf{753} & \textbf{1.6\%} & \textbf{753} & \textbf{1.6\%} & 759        & 2.4\%       & 761  & 2.7\%  & 758   & 2.3\%   \\
P-n55-k7 & 568      & 608& 7.0\% & \textbf{581} & \textbf{2.3\%} & 583& 2.6\%& 589        & 3.7\%       & \textbf{581}   & \textbf{2.3\%}   & \textbf{581}    & \textbf{2.3\%}    \\
P-n55-k10& 694      & 747& 7.6\% & 705& 1.6\%& 705& 1.6\%& 712        & 2.6\%       & 706  & 1.7\%  & \textbf{704}    & \textbf{1.4\%}    \\
P-n55-k15& 989      & 1047         & 5.9\% & 958& -3.1\%         & 974& -1.5\%         & 964        & -2.5\%      & 958  & -3.1\% & \textbf{948}    & \textbf{-4.1\%}   \\
P-n60-k10& 744      & 775& 4.1\% & 767& 3.1\%& 761& 2.3\%& 775        & 4.2\%       & 761  & 2.3\%  & \textbf{758}    & \textbf{1.9\%}    \\
P-n60-k15& 968      & 1026         & 5.9\% & 994& 2.7\%& 993& 2.6\%& 1002       & 3.5\%       & 990  & 2.3\%  & \textbf{984}    & \textbf{1.7\%}    \\
P-n65-k10& 792      & 809& 2.2\% & 813& 2.7\%& 808& 2.0\%& 827        & 4.4\%       & \textbf{803}   & \textbf{1.4\%}   & \textbf{803}    & \textbf{1.4\%}    \\
P-n70-k10& 827      & 863& 4.3\% & \textbf{844} & \textbf{2.1\%} & \textbf{844} & \textbf{2.1\%} & 856        & 3.5\%       & 846  & 2.3\%  & 846   & 2.3\%   \\
P-n76-k4 & 593      & 637& 7.4\% & \textbf{627} & \textbf{5.7\%} & 641& 8.1\%& 655        & 10.5\%      & 688  & 16.0\% & 643   & 8.4\%   \\
P-n76-k5 & 627      & 674& 7.5\% & \textbf{650} & \textbf{3.7\%} & 667& 6.4\%& 671        & 7.0\%       & 680  & 8.5\%  & 653   & 4.1\%   \\
P-n101-k4& 681      & 747& 9.7\% & \textbf{721} & \textbf{5.9\%} & 752& 10.4\%         & 751        & 10.3\%      & 764  & 12.2\% & 747   & 9.7\%   \\
\midrule
Average  &&    & 35.6\%&    & 4.2\%&    & 12.9\%         &  & 10.8\%      &      & 3.8\%  &       & \textbf{2.7\%}   \\
\bottomrule
\end{tabular}%
}
\caption{Results on CVRPLIB Set P instances.}~\label{table:setP}
\end{table*}

\begin{table*}[htbp]
\centering

\resizebox{0.9\textwidth}{!}{%
\begin{tabular}{lccccccccccccc}
\toprule
\multicolumn{1}{c}{Instance} & Baseline & \multicolumn{2}{c}{POMO}  & \multicolumn{2}{c}{POMO Multi}        & \multicolumn{2}{c}{Meta POMO}   & \multicolumn{2}{c}{Omni} & \multicolumn{2}{c}{Prompt (Ours)} & \multicolumn{2}{c}{Prompt top-8 (Ours)} \\
\midrule
X-n101-k25         & 27591    & 29381& 6.5\%& 28848& 4.6\%& 29232& 5.9\%& 29242       & 6.0\%      & 29350 & 6.4\% & \textbf{28397}   & \textbf{2.9\%}   \\
X-n106-k14         & 26362    & 27113& 2.9\%& \textbf{26684} & \textbf{1.2\%} & 26752& 1.5\%& 27005       & 2.4\%      & 27024 & 2.5\% & 26842  & 1.8\%  \\
X-n110-k13         & 14971    & 15235& 1.8\%& \textbf{15155} & \textbf{1.2\%} & 15397& 2.8\%& 15449       & 3.2\%      & 15286 & 2.1\% & 15167  & 1.3\%  \\
X-n115-k10         & 12747    & 13270& 4.1\%& 13702& 7.5\%& 13382& 5.0\%& 13573       & 6.5\%      & 13422 & 5.3\% & \textbf{13217}   & \textbf{3.7\%}   \\
X-n120-k6& 13332    & 13836& 3.8\%& 13650& 2.4\%& 14055& 5.4\%& 14006       & 5.1\%      & 13804 & 3.5\% & \textbf{13642}   & \textbf{2.3\%}   \\
X-n125-k30         & 55539    & 57958& 4.4\%& \textbf{57631} & \textbf{3.8\%} & 58489& 5.3\%& 58435       & 5.2\%      & 58585 & 5.5\% & 58143  & 4.7\%  \\
X-n129-k18         & 28940    & 29481& 1.9\%& 29544& 2.1\%& 29853& 3.2\%& 29920       & 3.4\%      & 29444 & 1.7\% & \textbf{29240}   & \textbf{1.0\%}   \\
X-n134-k13         & 10916    & 11378& 4.2\%& \textbf{11222} & \textbf{2.8\%} & 11291& 3.4\%& 11273       & 3.3\%      & 11353 & 4.0\% & 11229  & 2.9\%  \\
X-n139-k10         & 13590    & \textbf{13814} & \textbf{1.7\%} & 13961& 2.7\%& 14093& 3.7\%& 14007       & 3.1\%      & 13832 & 1.8\% & 13825  & 1.7\%  \\
X-n143-k7& 15700    & 16124& 2.7\%& 16288& 3.7\%& 16692& 6.3\%& 16597       & 5.7\%      & 16435 & 4.7\% & \textbf{16111}   & \textbf{2.6\%}   \\
X-n148-k46         & 43448    & 45673& 5.1\%& 45551& 4.8\%& 46409& 6.8\%& 46242       & 6.4\%      & 46080 & 6.1\% & \textbf{45374}   & \textbf{4.4\%}   \\
X-n153-k22         & 21220    & 23378& 10.2\%         & 23778& 12.1\%         & \textbf{22693} & \textbf{6.9\%} & 23317       & 9.9\%      & 23851 & 12.4\%& 23180  & 9.2\%  \\
X-n157-k13         & 16876    & 17805& 5.5\%& \textbf{17074} & \textbf{1.2\%} & 17340& 2.7\%& 17117       & 1.4\%      & 17182 & 1.8\% & 17182  & 1.8\%  \\
X-n162-k11         & 14138    & 14732& 4.2\%& 14601& 3.3\%& 14843& 5.0\%& 14660       & 3.7\%      & 14485 & 2.5\% & \textbf{14473}   & \textbf{2.4\%}   \\
X-n167-k10         & 20557    & 21398& 4.1\%& 21224& 3.2\%& 21634& 5.2\%& 21512       & 4.6\%      & \textbf{21014}  & \textbf{2.2\%}  & \textbf{21014}   & \textbf{2.2\%}   \\
X-n172-k51         & 45607    & 48775& 6.9\%& 48174& 5.6\%& 48746& 6.9\%& 48604       & 6.6\%      & 49241 & 8.0\% & \textbf{47904}   & \textbf{5.0\%}   \\
X-n176-k26         & 47812    & 52214& 9.2\%& 52651& 10.1\%         & \textbf{50642} & \textbf{5.9\%} & 52368       & 9.5\%      & 52843 & 10.5\%& 52194  & 9.2\%  \\
X-n181-k23         & 25569    & 28243& 10.5\%         & \textbf{26020} & \textbf{1.8\%} & 26371& 3.1\%& 26046       & 1.9\%      & 26145 & 2.3\% & 26103  & 2.1\%  \\
X-n186-k15         & 24145    & 25540& 5.8\%& 24889& 3.1\%& 25451& 5.4\%& 24898       & 3.1\%      & \textbf{24673}  & \textbf{2.2\%}  & \textbf{24673}   & \textbf{2.2\%}   \\
X-n190-k8& 16980    & 18359& 8.1\%& \textbf{17341} & \textbf{2.1\%} & 17869& 5.2\%& 17744       & 4.5\%      & 17717 & 4.3\% & 17635  & 3.9\%  \\
X-n195-k51         & 44225    & 48297& 9.2\%& \textbf{46818} & \textbf{5.9\%} & 47391& 7.2\%& 48079       & 8.7\%      & 47817 & 8.1\% & 46864  & 6.0\%  \\
X-n200-k36         & 58578    & 61926& 5.7\%& \textbf{61025} & \textbf{4.2\%} & 61385& 4.8\%& 61150       & 4.4\%      & 61779 & 5.5\% & 61236  & 4.5\%  \\
\midrule
Average  &&      & 5.4\%&      & 4.1\%&      & 4.9\%&   & 4.9\%      &       & 4.7\% &        & \textbf{3.5\%}  \\
\bottomrule
\end{tabular}%
}
\caption{Results on CVRPLIB Set X 100-200 instances.}~\label{table:setX}
\end{table*}

\begin{table*}[htbp]
\centering

\resizebox{0.9\textwidth}{!}{%
\begin{tabular}{lccccccccccccc}
\toprule
\multicolumn{1}{c}{Instance} & Baseline & \multicolumn{2}{c}{POMO}  & \multicolumn{2}{c}{POMO Multi}        & \multicolumn{2}{c}{Meta POMO} & \multicolumn{2}{c}{Omni}        & \multicolumn{2}{c}{Prompt (Ours)} & \multicolumn{2}{c}{Prompt top-8 (Ours)} \\
\midrule
XML100\_1116\_13   & 9528     & 10399& 9.1\%& 10263& 7.7\%& 10579         & 11.0\%        & 10693& 12.2\%         & 10334 & 8.5\% & \textbf{10208}   & \textbf{7.1\%}   \\
XML100\_1124\_25   & 11515    & 11848& 2.9\%& 11944& 3.7\%& 12203         & 6.0\%         & 12274& 6.6\%& 12488 & 8.4\% & \textbf{11799}   & \textbf{2.5\%}   \\
XML100\_1154\_14   & 16123    & 16391& 1.7\%& 16588& 2.9\%& 16642         & 3.2\%         & 16646& 3.2\%& 16543 & 2.6\% & \textbf{16374}   & \textbf{1.6\%}   \\
XML100\_1163\_08   & 12863    & 13142& 2.2\%& 13103& 1.9\%& 13401         & 4.2\%         & 13386& 4.1\%& 13137 & 2.1\% & \textbf{13094}   & \textbf{1.8\%}   \\
XML100\_1215\_26   & 4983     & 5415 & 8.7\%& \textbf{5093}  & \textbf{2.2\%} & 5412& 8.6\%         & 5309 & 6.5\%& 5405  & 8.5\% & 5195   & 4.3\%  \\
XML100\_1311\_26   & 29707    & 31196& 5.0\%& \textbf{29905} & \textbf{0.7\%} & 30292         & 2.0\%         & 30593& 3.0\%& 30156 & 1.5\% & 30130  & 1.4\%  \\
XML100\_1334\_17   & 11559    & 11798& 2.1\%& 11802& 2.1\%& 12072         & 4.4\%         & 12009& 3.9\%& 12288 & 6.3\% & \textbf{11655}   & \textbf{0.8\%}   \\
XML100\_1372\_07   & 20201    & 21983& 8.8\%& 21712& 7.5\%& 21700         & 7.4\%         & 21608& 7.0\%& 21695 & 7.4\% & \textbf{21275}   & \textbf{5.3\%}   \\
XML100\_2125\_05   & 10449    & 10994& 5.2\%& \textbf{10922} & \textbf{4.5\%} & 11103         & 6.3\%         & 11326& 8.4\%& 11397 & 9.1\% & 10950  & 4.8\%  \\
XML100\_2165\_03   & 9053     & \textbf{9546}  & \textbf{5.4\%} & 9593 & 6.0\%& 9719& 7.4\%         & 9855 & 8.9\%& 10215 & 12.8\%& 9557   & 5.6\%  \\
XML100\_2176\_24   & 8970     & \textbf{9580}  & \textbf{6.8\%} & 9673 & 7.8\%& 9921& 10.6\%        & 10258& 14.4\%         & 10327 & 15.1\%& 9601   & 7.0\%  \\
XML100\_2223\_26   & 12992    & 13383& 3.0\%& 13352& 2.8\%& 13462         & 3.6\%         & 13370& 2.9\%& 13429 & 3.4\% & \textbf{13349}   & \textbf{2.7\%}   \\
XML100\_2275\_04   & 8435     & 8782 & 4.1\%& 8817 & 4.5\%& 8946& 6.1\%         & 8975 & 6.4\%& 8830  & 4.7\% & \textbf{8779}    & \textbf{4.1\%}   \\
XML100\_2364\_08   & 9797     & 10258& 4.7\%& \textbf{10133} & \textbf{3.4\%} & 10417         & 6.3\%         & 10450& 6.7\%& 10581 & 8.0\% & 10230  & 4.4\%  \\
XML100\_3123\_14   & 23616    & 23940& 1.4\%& 24081& 2.0\%& 24455         & 3.6\%         & 24226& 2.6\%& 24011 & 1.7\% & \textbf{23903}   & \textbf{1.2\%}   \\
XML100\_3165\_17   & 14116    & \textbf{14387} & \textbf{1.9\%} & 14541& 3.0\%& 14810         & 4.9\%         & 14884& 5.4\%& 14692 & 4.1\% & 14405  & 2.0\%  \\
XML100\_3215\_08   & 12912    & 13289& 2.9\%& \textbf{13129} & \textbf{1.7\%} & 13283         & 2.9\%         & 13470& 4.3\%& 13467 & 4.3\% & 13231  & 2.5\%  \\
XML100\_3243\_04   & 17654    & 18185& 3.0\%& \textbf{17830} & \textbf{1.0\%} & 17957         & 1.7\%         & 18007& 2.0\%& 18000 & 2.0\% & 17933  & 1.6\%  \\
XML100\_3254\_22   & 14865    & 15439& 3.9\%& 15199& 2.2\%& 15260         & 2.7\%         & 15610& 5.0\%& 15536 & 4.5\% & \textbf{15185}   & \textbf{2.2\%}   \\
XML100\_3371\_20   & 55824    & 59145& 5.9\%& 58868& 5.5\%& 58251         & 4.3\%         & \textbf{57723} & \textbf{3.4\%} & 59483 & 6.6\% & 58470  & 4.7\%  \\
\midrule
Average  &&      & 4.4\%&      & 3.7\%&     & 5.4\%         &      & 5.8\%&       & 6.1\% &        & \textbf{3.4\%}\\
\bottomrule
\end{tabular}%
}
\caption{Results on XML instances.}~\label{table:setXML}
\end{table*}

\end{document}